\documentclass[a4paper,conference]{IEEEtran}

\usepackage{microtype}
\usepackage{graphicx}
\usepackage{subfigure}
\usepackage{booktabs} 
\usepackage[table]{xcolor}

\usepackage{hyperref}

\newcommand{\lt}[1]{_{\text{#1}}}

\usepackage{algorithm,algorithmic,amsmath}

\usepackage[utf8]{inputenc}

\begin{document}

\title{Iterative Label Improvement: Robust Training by Confidence Based Filtering and Dataset Partitioning}

\author{\IEEEauthorblockN{Christian Haase-Sch\"utz\IEEEauthorrefmark{1}\IEEEauthorrefmark{2}, Rainer Stal\IEEEauthorrefmark{1}, Heinz Hertlein\IEEEauthorrefmark{1}, Bernhard Sick\IEEEauthorrefmark{2}}
	\IEEEauthorblockA{\IEEEauthorrefmark{1}Engineering Cognitive Systems, Automated Driving, Robert Bosch GmbH, Abstatt, Germany
		\IEEEauthorblockA{\IEEEauthorrefmark{2}Intelligent Embedded Systems Group, University of Kassel, Kassel, Germany}}
	\IEEEauthorblockA{\{christian.schuetz2, rainer.stal, heinz.hertlein\}@de.bosch.com, bsick@uni-kassel.de}}

\maketitle



\begin{abstract}
State-of-the-art, high capacity deep neural networks not only require large amounts of labelled training data, they are also highly susceptible to labelling errors in this data, typically resulting in large efforts and
costs and therefore limiting the applicability of deep learning. To alleviate this issue, we propose a novel meta training and labelling scheme that is able to use inexpensive unlabelled data
by taking advantage of the generalization power of deep neural networks.
We show experimentally
that by solely relying on one network architecture and our
proposed scheme of combining self-training with pseudo-labels, both label quality and resulting model accuracy, can be
improved significantly.
Our method achieves state-of-the-art results, while being architecture agnostic and therefore broadly applicable. 
Compared to other methods dealing with erroneous labels, our approach does neither require another network to be trained,
nor does it necessarily need an additional, highly accurate reference label set.
Instead of removing samples from a labelled set,
our technique uses additional sensor data without the need for manual labelling. Furthermore, our approach can be used for semi-supervised learning.
\end{abstract}

\section{Introduction}

	Supervised deep learning methods deliver state-of-the-art results in many important applications, reaching or in some cases
	even surpassing human-level performance. As a result, deep learning has been adopted for a variety of challenging pattern recognition tasks, such as medical image classification \cite{taghanaki2019deep, lundervold2019overview}, power forecasting for renewable energy plants \cite{gensler2016deep}, or autonomous driving \cite{feng2019deep}.

	\begin{figure}[tbp]
		\centering
		\centering
		\includegraphics[width=0.85\linewidth]{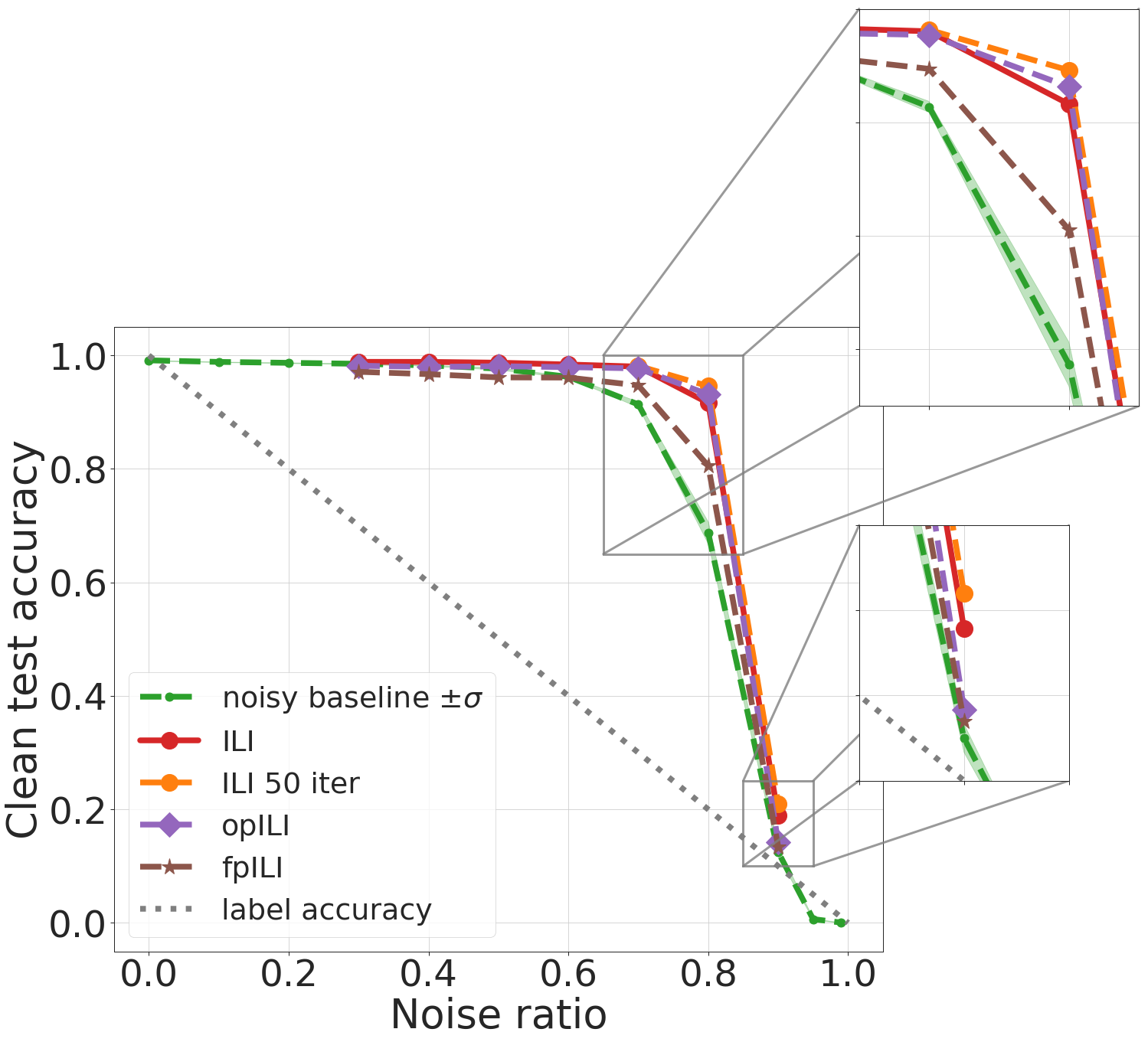}
		\caption{Comparison of different versions of our ILI algorithm on MNIST with erroneous labels. All versions of our method improve the accuracy on the test set, as compared to the noisy baseline. For reference the empirical one $\sigma$ interval is shown for the baseline.}\label{fig:acc_noise_mnist}
		\vspace{-2.5mm}
	\end{figure}
	A major drawback deep neural networks suffer from, especially in supervised learning, is the necessity for a large amount of accurately labelled training data, if good performance is to be achieved. 
	This can prevent or delay the utilisation of deep learning based techniques for practical applications
	due to the large effort and high cost of producing labels, e.g. by manual labelling.
	Therefore, it is desirable to make effective use of unlabelled data and increase the robustness of the learning procedure against incomplete and erroneous labels.
    As a step towards this goal, we propose
	a novel, iterative technique to train deep neural networks, making use of unlabelled training data
	in addition to a small or partially wrongly labelled initialisation data set. In order to demonstrate the feasibility of the approach,
	a comprehensive study of the effects of label noise (fraction of errors) in the training material in a classification setting is conducted. 
	
	Both training progress and training outcome in terms of the error rates on unseen data are related to
	the closely interlinked factors of the chosen cost function, uncertainty of the prediction outputs and labelling noise.
	High cost during training means the prediction from the network and the label presented differ to a large extent.
	The reason can be twofold, first, it could mean that the network has not yet learned the sample presented so 
	it is highly informative and hence should be considered as very relevant for the training progress; second, it could mean the sample is not labelled correctly, in which case it should be ignored.

	In spite of this difficulty, we show the feasibility of an iterative technique to improve the label quality automatically, requiring neither manual intervention nor carefully handcrafted filtering algorithms reducing the amount of training material needed.
	Additionally, our approach allows to leverage automated label correction in order to exploit initially unlabelled data and as a result to improve
	the training outcome and recognition performance, making it suitable for semi-supervised learning (SSL).

\section{The effects of erroneous labels}\label{sec:effect}
Erroneous labels are a well studied problem which is especially relevant for practical applications. Even for the classification of bio-medical data, where data is particularly precious, labelling erros occur \cite{alon1999broad, li2001gene, zhang2009methods} and are studied further in classical machine learning settings, e.g., in \cite{malossini2006detecting, bootkrajang2012label}. 
Recent works study the effect of erroneous labels on object detection for automated driving using deep neural networks \cite{chadwick2019training} and \cite{haase2019estimating}.
Deep neural networks are known to have the capability of learning arbitrary assignments of labels to samples, provided that the model capacity is high enough.
As a result, labelling errors in the training set typically cause a large difference between training and validation loss \cite{zhang2016understanding}.
Consequently, an insufficient amount of correctly labelled training data leads to similar effects, as the model can adapt perfectly
to the training set, but the result on unseen data may be unsatisfactory.
Multiple methods that deal with erroneous labels in the training material for deep neural networks have been proposed, e.g. \cite{ren2018learning, jiang2017mentornet, nguyen2019self, nguyen2019robust}.
 \\ 
Erroneous labels or an insufficient amount of well-labelled training data
occur when real-world problems have to be solved under strict timely and monetary constraints; or when collected and labelled
data is extremely valuable, such as for medical applications, where additional experiments with diseases are ethically impossible or label correction prohibited due data privacy. The
resulting labels might not be completely wrong - rather inaccurate, and the data sample highly valuable, hence it is desirable to make the best possible use of them instead
of filtering them out completely. Such errors could either be due to some bias, e.g. samples of class \textit{A} typically falsely labelled as \textit{B}, or randomly distributed inaccuracies. 

\section{Related work and the idea of Iterative Label Improvement (ILI)}
\begin{figure}[tbp]
	\centering
	\includegraphics[width=0.9\linewidth]{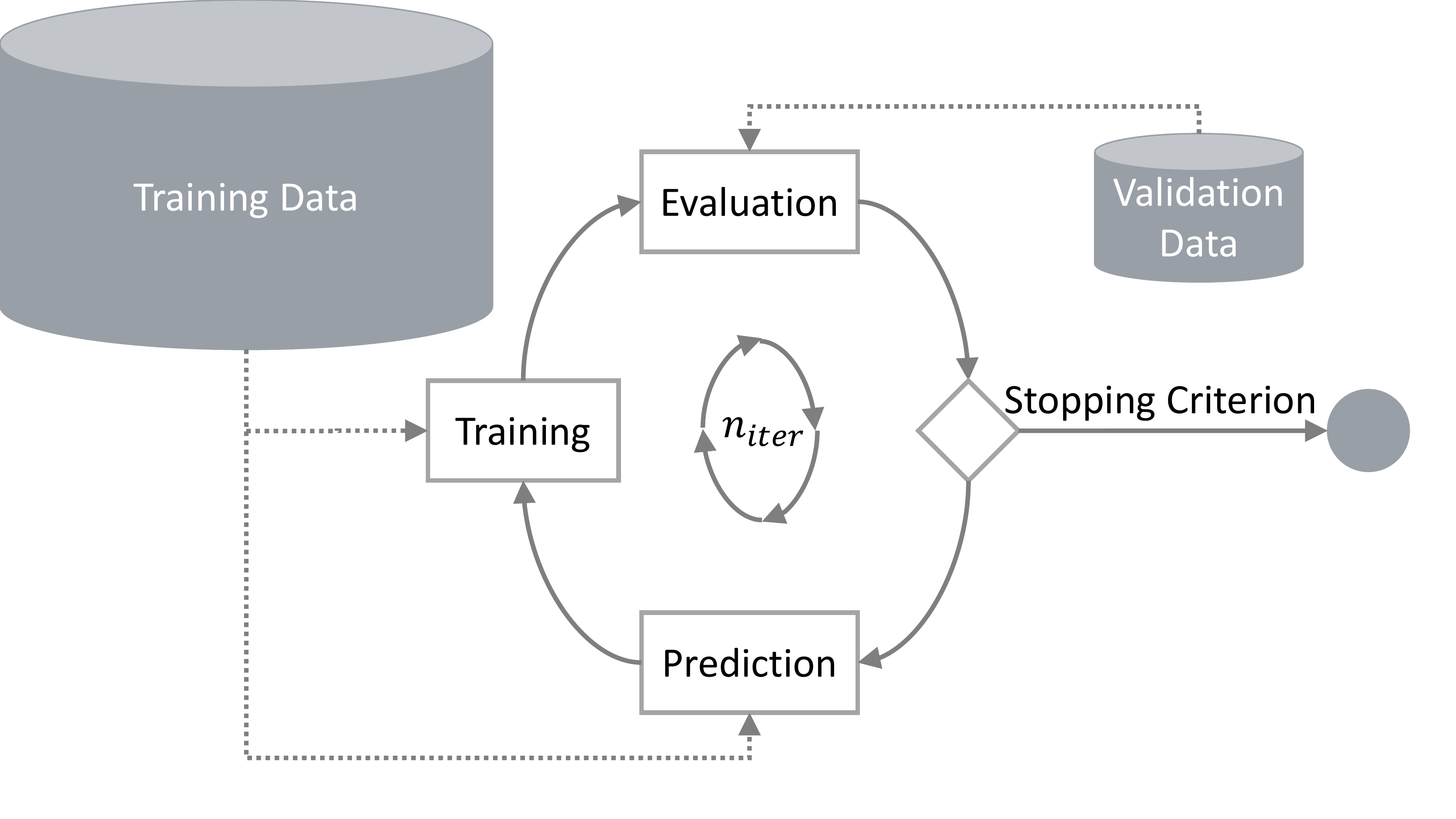}
	\caption{Scheme of our proposed ILI approach. By suitably choosing the data to train and predict on, and using the standard train, predict, and evaluate scheme, ILI is able to improve the labelling quality. Training data is assumed to contain erroneous labels. The test needs to be clean, i.e. well-labelled and is used to determine the performance of the algorithm. If the validation data is used to determine an early stopping point, this dataset can also contain errors. Depending on the ILI variant used, the erroneous training data can contain a small amount of reference labels or additional unlabelled training data, the fraction of labelled training data can vary.}\label{fig:algo_scheme}
	\vspace{-2.5mm}
\end{figure}

Inspired by the classical training and predicting scheme and leveraging the generalisation power of deep neural networks, we employ an iterative scheme as depicted in Figure \ref{fig:algo_scheme}, using the predictions of a trained network as training material for another instance of such a network. The trained network provides so-called pseudo-labels for previously unlabelled data. 

Other approaches dealing with label noise in the training material for deep neural networks are based on finding an optimal curriculum, to present the data, as first introduced in \cite{bengio2009curriculum}. This approach has the potential to make the best possible use of the given labels. 
Traditional curricula, however, require careful engineering, which can limit practical applicability of the technique.
MentorNet \cite{jiang2017mentornet} attempts to learn a curriculum from data, requiring an additional teacher network to be trained jointly with the student net, i.e., the network learning the actual problem.
Learning to reweight \cite{ren2018learning} on the other hand employs a clean validation set, requiring the labels to be of high quality in order to be able to judge the quality of the training data. 
Instead of using curricula, other methods such as \cite{nguyen2019self} and \cite{nguyen2019robust} rely on filtering the noisy labels,
i.e. getting rid of them. Instead, our approach estimates improved labels for potentially noisy samples in order to gain advantage of the additional training material.
Although curriculum learning, with a suitably chosen curriculum, limits the adverse effect of badly labelled samples on the
training and the resulting model, it does not take advantage of automatic label generation as our approach. Therefore,
iterative label improvement can take advantage of additional, initially unlabelled data which is usually inexpensively available in large quantities
in many practical cases. Making it a technique suitable for SSL.
While filtering out bad samples might be desirable to increase a model's accuracy, in settings with highly valuable samples, it is beneficial to improve the existing labels, as is possible with our proposed approach.

Self-training, i.e. training a network using its predictions \cite{scudder1965probability}, and pseudo-labels are employed by \cite{xie2020self} and \cite{yalniz2019billion}, in a SSL setting. In contrast to those works we provide an iterative mechanism for label improvement and rely on a single network architecure. The idea of iteratively applying pseudo-labels has been applied in iterative cross learning (ICL) \cite{yuan2018iterative}. Our approach extends the basic idea presented there by using a confidence filter, and taking full advantage of multiple optimisation iterations. We formulate our approach such that it can be used for SSL. We further study the influence of data augmentation, which is often neglected in studies on erroneous labels, despite being a standard technique in computer vision.

Let $m$ be a model, which for a sample $x$ in the training set $X\lt{train}$ predicts a target $\hat{y}$ with a confidence $c$ and $\mathcal{F}$ a filter function, deciding between label versions $y$ and $\hat{y}$. The model is initially trained on label $y\lt{train}$ and $(i)$ indicates the number of the current iteration, $n_{\text{iter}}$ in total.
The novel technique we propose is referred to as ``Iterative Label Improvement" or ILI.\footnote{The code for ILI will be made public after the paper has been accepted for publication.}
 There are several related variants of the approach
which will be described in the following sections. In algorithm \ref{algo:plainIli_initIli}, the first simple variant is shown as an example (for details see Section \ref{subsec:algo:plain}). The final model and the improved labels can be obtained from the last ILI iteration. \\

\begin{algorithm}
	\caption{plainILI filtered with initILI}
	\label{algo:plainIli_initIli}
	\begin{algorithmic}[1]
		\renewcommand{\algorithmicrequire}{\textbf{Input:}}
		\renewcommand{\algorithmicensure}{\textbf{Output:}}
		\REQUIRE $X_{\text{train}}$, $y_{\text{train}}$, $n_{\text{iter}}$
		\ENSURE  Model $m^{n_{\text{iter}}}$, $y^{n_{\text{iter}}}_{\text{train}}$

		\STATE $m^{(0)} = m.initialize()$
		\STATE $m^{(0)}.fit(X_{\text{train}}, y_{\text{train}})$

		\FOR {$i = 1$ to $n_{\text{iter}}$}

		\STATE $y_{\text{train}}^{(i)} = \mathcal{F}\left[m^{(i-1)}\right]\left(X_{\text{train}}, y_{\text{train}}^{(i-1)}\right)$ \\
		\STATE $m^{(i)} = m.initialize()$ \\
		\STATE $m^{(i)}.fit(X_{\text{train}}, y_{\text{train}}^{(i)})$
		
		\ENDFOR
	\end{algorithmic} 
\end{algorithm}

\section{The ILI approach in detail}
We train our model using a student-teacher paradigm, with a fixed architecture. Each iteration can be considered self-training. In each iteration we compare the pseudo-labels, i.e. predictions, with the given labels (if any) and decide using a filter function, which label is to be kept. 

In contrast to ICL, we use more iterations and rely on one network per iteration only. Our approach can be extended to multiple partitions, and we will show how to employ it for SSL. We use the certainty of the predictions represented by the confidence of the model, to filter the pseudo-labels. 

Fig. \ref{fig:algo_scheme} illustrates our proposed approach. Depending on the variant of the approach as described below,
the training data may contain labelling noise or a well-labelled, but small reference training set. In the latter case,
unlabelled data must be available additionally. If labelling noise is present, such labels will be called erroneous.

Regardless of the ILI variant, the separate test set does not contain erroneously labelled samples and its main
use is to evaluate the trained model in order to verify the performance of our proposed technique,
for example for the experimental results presented in this paper. In a practical application,
other ways to evaluate the performance can be applied such as manual inspection of a selection of the labels generated. The validation set used to find a point for early stopping can contain errors as well.

There are several variants of the ILI scheme which differ in the following three aspects: (i) number of partitions, the unlabelled data is split into; (ii) quality of the initialisation dataset; and (iii) label selection or filtering approach.
When deploying ILI to an application, the respective variants of these three aspects can be combined as suited best for the particular case.
For all algorithms, we initialize the model in each optimisation step, i.e. the model is trained from scratch.

\subsection{Number of partitions}
ILI can be applied using the complete training dataset at once (no partitioning) or using various splits (with partitioning). 

\subsubsection{plainILI: No partitioning}\label{subsec:algo:plain}

A simple variant of the approach which is referred to as ``plainILI" is to always use the same set of samples for training and prediction,
as in algorithm \ref{algo:plainIli_initIli}.
This makes sense if an automated technique for generating inexpensive, but faulty initialisation labels for the first iteration is available.
The model is trained on the labels in spite of the errors. In the next step, we apply the trained model on the samples of the training set,
ignoring the existing labels and then replacing all of them with the predictions.
So the plain algorithm replaces the labels in each ILI iteration with the predictions of the network from the previous ILI iteration, during which it was trained a predefined number of epochs,
so in iteration $i$, $y\lt{train}^{(i)} = \text{argmax}(m^{(i-1)}.\text{predict}(X\lt{train}))$, which is to be understood per sample. In each iteration, the model is trained from scratch to avoid overfitting to the noisy training labels.

\subsubsection{Partitioning based Iterative Label Improvement -- pILI}\label{subsec:algo:pili}
Partitioning based ILI (pILI) employs a splitting of the training data into at least two parts, in order to reduce the effect that the trained model
will learn the errors in the labels, and, therefore to improve generalisation.
Instead of using all training data at once, pILI alternates between training on a subset $\mathcal{A}$ of the data and predicting labels on a different
subset $\mathcal{B}$ (details on the choice of subsets given below), for each ILI iteration.
So, the predictions made by a model are always on samples that particular model has not seen in training.
To prevent error propagation as much as possible, the training process always starts from scratch in each optimisation iteration,
so the reinitialization of the model results in not using any history that would otherwise be contained in the model's weights. There are two variants of pILI, oscillating partitioning (opILI) and fraction partitioning ILI (fpILI).

\paragraph{opILI: Oscillation based ILI}
Exactly two subsets $\mathcal{A}$ and $\mathcal{B}$ are used alternately,
so training and prediction oscillate between these two disjoint subsets.
Regarding the initial labels needed to start the process, one variant is shown in algorithm \ref{algo:opIliWithInitIli}.
In this case, initialisation labels are needed only for $\mathcal{A}$, making this variant especially applicable to SSL, and these labels may contain errors, as they are used only at the beginning of the process.


\begin{algorithm}
	\caption{opILI with initILI}
	\label{algo:opIliWithInitIli}
	\begin{algorithmic}[1]
		\renewcommand{\algorithmicrequire}{\textbf{Input:}}
		\renewcommand{\algorithmicensure}{\textbf{Output:}}
		\REQUIRE $X_{\text{train, A}}$, $y_{\text{train}}$, $n_{\text{iter}}$
		\ENSURE  Model $m^{n_{\text{iter}}}$, $y^{n_{\text{iter}}}_{\text{train}}$
		
		\STATE $X\lt{train, A}, X\lt{train, B} = \text{Split}(X\lt{train})$
		
		\STATE $m^{(0)}\lt{A}.initialize()$
		\STATE $m^{(0)}\lt{A}.fit(X\lt{train, A}, y^{(0)}\lt{train, A})$
		\FOR {$i = 1$ to $n\lt{iter}$}
			\STATE $y\lt{train, B}^{(i)} = m^{(i-1)}\lt{A}\left(X\lt{train, B}\right)$ \\
			\STATE $m^{(i)}\lt{B}.initialize()$ \\
			\STATE $m^{(i)}\lt{B}.fit(X\lt{train, B}, y\lt{train, B}^{(i)})$ \\
			
			\STATE $y\lt{train, A}^{(i)} = m^{(i)}\lt{B}\left(X\lt{train, A}\right)$ \\
			\STATE $m^{(i)}\lt{A}.initialize()$ \\
			\STATE $m^{(i)}\lt{A}.fit(X\lt{train, A}, y\lt{train, A}^{(i)})$ \\

		\ENDFOR
		\STATE $y\lt{train}^{n_{\text{iter}}} = \left[y\lt{train, A}^{(n_{\text{iter}})}, y\lt{train, B}^{(n_{\text{iter}})}\right]$	\end{algorithmic} 
\end{algorithm}

\paragraph{fpILI: Fragmentation based ILI}

The training data is split into multiple subsets, a labelled set $X\lt{train, A}$ and multiple unlabelled sets $X\lt{train, B$_i$} \quad i=0,...,n$,
where $n>1$ is a hyperparameter. All partitions in $\mathcal{B}$ are usually similar in size, i.e., $\vert X\lt{train, B$_i$}\vert \approx \vert X\lt{train, B$_k$} \vert ~\text{for all}~i, k: 0 \leq i, k \leq n$.
Only the initialisation set $X\lt{train, A}$  is labelled with $y\lt{train, A}$.
The subsets are usually (but not necessarily) a partition of the overall training set in the mathematical sense:
\[X\lt{train} = X\lt{train, A} \cup \left(\bigcup_i X\lt{train, B$_i$}\right).\]

The advantage of fpILI in comparison to opILI is that in each new optimisation iteration,
the samples for which the trained model makes predictions are unseen not only with regards to the training
data of the model of ILI iteration $(i)$, but also with regards to the labels of the training set of iteration $(i)$, which are influenced implicitly by the data from the previous iterations $(0,...,i-1)$.

For this reason, fpILI should in principle be preferred over opILI, however more unlabelled data is required and is particularly useful for SSL.
If the availability of unlabelled data is restricted, opILI may perform better in practice.

\subsection{Quality of initialisation data set}
Depending on the quality of the initial dataset, different versions of ILI can be used, which will be introduced in the following. 

\subsubsection{initILI: initial data only used once}

In both algorithms \ref{algo:plainIli_initIli} and \ref{algo:opIliWithInitIli}, it is acceptable if
the initialisation data set contains faulty labels, as in both cases, the initial labels are used
only in the first optimisation iteration. This variant is called initILI, as the given labels
are used for initialisation only.

\subsubsection{refILI: high quality reference data}
The principle of ILI, however, can be useful as well if a data set $X\lt{train, A}$
with virtually perfect labels is available yet the size of this typically manually labelled
reference set is too small to reach the needed recognition performance.
In this case, the reference set $X\lt{train, A}$ is used to train an initial model to make
predictions on a larger set in the first ILI iteration. In each subsequent training,
both the predicted labels from the previous iteration and the reference set is used together
for training. This ILI variant is called refILI. Regarding the additional, initially
unlabelled set $X\lt{train, B}$, depending on how many partitions are created,
refILI is combined with opILI (exactly two partitions of $X\lt{train, B}$) or fpILI
(more than two partitions).

Algorithm \ref{algo:FpiliWithRefIli} shows as an example refILI combined with fpILI.
Considering that the labels of $X\lt{train, A}$ are more accurate than the predicted labels
of $X\lt{train, B}$ (especially at the beginning of the ILI process) and  $X\lt{train, A}$
is smaller than each one of the single partitions $X\lt{train, B$_i$}$, an approach to increase
the relative influence of the samples in $X\lt{train, A}$ should be adopted.

\begin{algorithm}
	\caption{refILI with fpILI}
	\label{algo:FpiliWithRefIli}
	\begin{algorithmic}[1]
		\renewcommand{\algorithmicrequire}{\textbf{Input:}}
		\renewcommand{\algorithmicensure}{\textbf{Output:}}
		\REQUIRE $X\lt{train}$, $y\lt{train}$, $n\lt{iter}$
		\ENSURE  Model $m^{n_{\text{iter}}}$, $y^{n_{\text{iter}}}_{\text{train}}$
		
		\STATE $X\lt{train, A}, X\lt{train, B$_0$}, ..., X\lt{train, B$_n$}  = \text{Split}(X\lt{train})$
		
		\STATE $m\lt{A}.initialize()$
		\STATE $m\lt{A}.fit(X\lt{train, A}, y\lt{train, A})$
		\STATE $y\lt{train, B$_0$} = m\lt{A}\left(X\lt{train, B$_0$}\right)$ \\
		
		\FOR {$i = 1$ to $n$}
		
		\STATE $m^{(i)}\lt{AB}.initialize()$ \\
		\STATE $m^{(i)}\lt{AB}.fit \left(\left[X\lt{train, A}, X\lt{train, B$_{i-1}$}\right], \left[y\lt{train, A}, y\lt{train, B$_{i-1}$}\right] \right)$ \\
		
		\STATE $y\lt{train, B$_i$} = m^{(i)}\lt{AB}\left(X\lt{train, B$_i$}\right)$ \\
		
		\ENDFOR
		\STATE $y^{(n_{\text{iter}})}\lt{train} = \left[y\lt{train, A}, y^{(n_{\text{iter}})}\lt{train, B$_{i-1}$}, ..., y^{(n_{\text{iter}})}\lt{train, B$_{i-1}$}\right]$
	\end{algorithmic} 
\end{algorithm}

\subsection{Label selection or filtering}
In contrast to pure self-training, we consider filtering labels based on the models uncertainty. Depending on the model of the current iteration, a selection or filtering function can be applied, to decide whether to use the labels of the previous or the current iteration. The different versions are introduced in the following.
This approach is loosely inspired by the confidence based acquisition function used in active learning \cite{settles2009active}.

\subsubsection{ILI unfiltered}
 Is the most basic approach, where the predictions of the model are kept as labels of the present iteration. After each iteration $i-1$, for the next iteration we set $y^{(i)} \overset{set}{=} \hat{y}^{(i-1)} = m^{(i-1)}(x)$, i.e. we trust the current models' predictions on the training data $x$.

\subsubsection{ILI filtered}
To account for model uncertainty, a filter function, which is a functional of the model built in the current iteration can be applied. For each iteration $i$, we use a filter, $\mathcal{F}$ a functional of the model $m^{(i-1)}$:  $y^{(i)} \overset{set}{=} \mathcal{F}^{(i-1)}(y^{(i-1)}) := \mathcal{F}\left[m^{(i-1)}\right](y^{(i-1)})$. For example, if using the confidence $c^{(i)}$ given by the model

\begin{equation*}
\begin{aligned}
\mathcal{F}^{(i)}_{\text{confidence}}\left(y_{\text{train}}^{(i)}\right) = 
\begin{cases}
\hat{y}_{\text{train}, i} = m^{(i)}(x_{\text{train}, i}): \quad &c^{(i)} > \vartheta \\
\hat{y}_{\text{train}, i} = y_{\text{train}, i}: \quad &\text{else.} 
\end{cases}
\end{aligned}
\end{equation*}

The filter function $\mathcal{F}$ is used to determine which labels of the current optimisation iteration step
are kept unchanged and which ones are replaced by the models' predictions.
This filtering adopts a suitable metric and typically relies on a threshold decision, accepting a particular prediction of a model if
the metric reaches or exceeds the threshold, while otherwise the original label is kept instead.
The metric can be defined, for example, as a measure of the uncertainty of the model's prediction,
which can be obtained from ensemble variation \cite{freeman1965elementary}, by Monte-Carlo Dropout \cite{gal2016dropout}
or simply from the confidence of the predictions, i.e. the value of the softmax in the last layer of the neural network for the winning class.

\section{Experiments}\label{sec:exps}
A prerequisite for our proposed ILI approach to be successful is the capability of deep neural networks to predict labels with
an error rate that is smaller than the error rate of the labels in the training data and, crucially,
to what extent this capability can be sustained in an iterative process. This determines
the overall effectiveness of ILI in its different variants in terms of the eventual recognition accuracy
that is achievable, starting from a certain quality of the labelled initialisation data set at the beginning. In settings where the amount of errors in the labels is so large, that the trained network can not generalise sufficiently, we do not expect our approach to lead to any improvement, without an additional regularisation technique.
It can be noted that this question, i.e., how effective ILI is in reducing labelling errors, is critical for all the ILI variants including refILI, even though a well labelled reference training set is available in this case. This is because for the deployment of refILI in practice the size of the nearly perfectly labelled reference data set is expected to be insufficient\footnote{In practice such a reference data set can be realized by spending large amounts of time and money per sample, hence it is desirable to limit its size.}.
As a result, the first training with the reference set alone will result in a model which obtains a recognition rate on unseen data significantly worse than the high accuracy of the reference set. Therefore, the success of ILI in this case depends on the capability to reduce the errors of predicted labels in an iterative fashion, similar to what is needed for initILI, when no reference set is available.
Keeping this in mind, refILI will not be included explicitly in the following
experimental results. Instead, the shown experiments are considered to be equally
relevant for both initILI and refILI.

\begin{figure*}[tpb]
	\centering
	\begin{minipage}{0.99\linewidth}
		\centering
		\subfigure[CIFAR-CNN on noisy MNIST data with random error. ]{\label{fig:mnist_random_cifar}\includegraphics[width=0.24\textwidth]{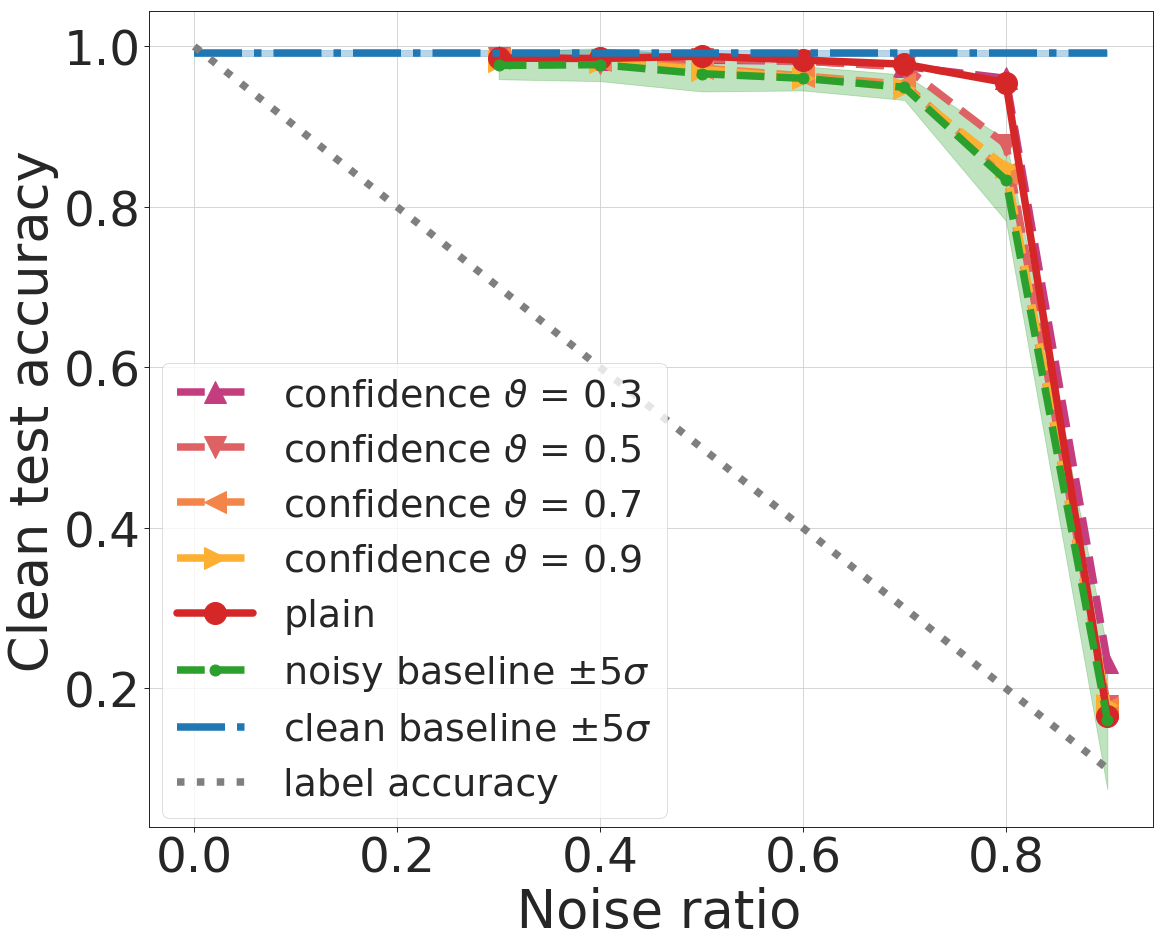}}
		\hfill
		\subfigure[MNIST-CNN on noisy MNIST data with random error. ]{\label{fig:mnist_random_mnist}\includegraphics[width=0.24\textwidth]{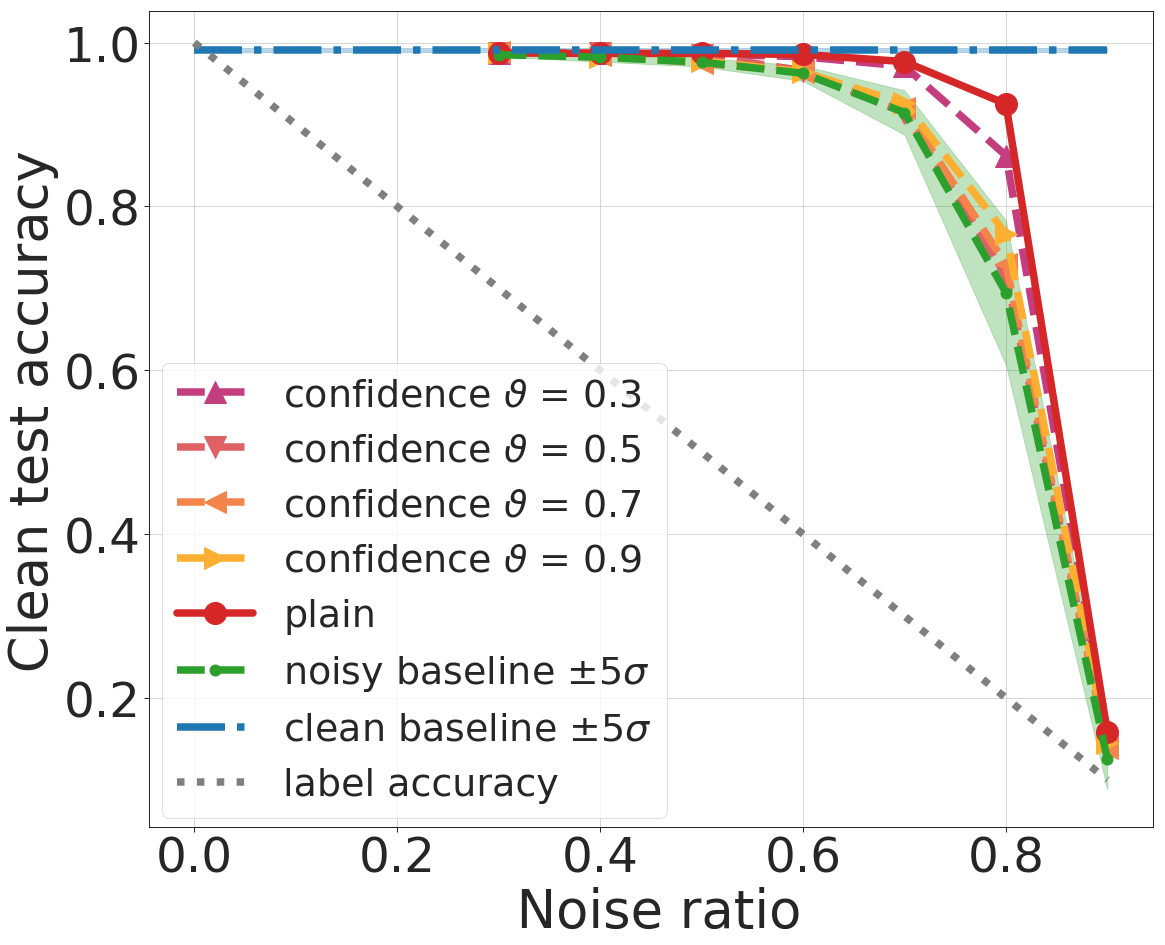}}
		\hfill
		\centering
		\subfigure[CIFAR-CNN on noisy MNIST data with bias error. ]{\label{fig:mnist_bias_cifar}\includegraphics[width=0.24\textwidth]{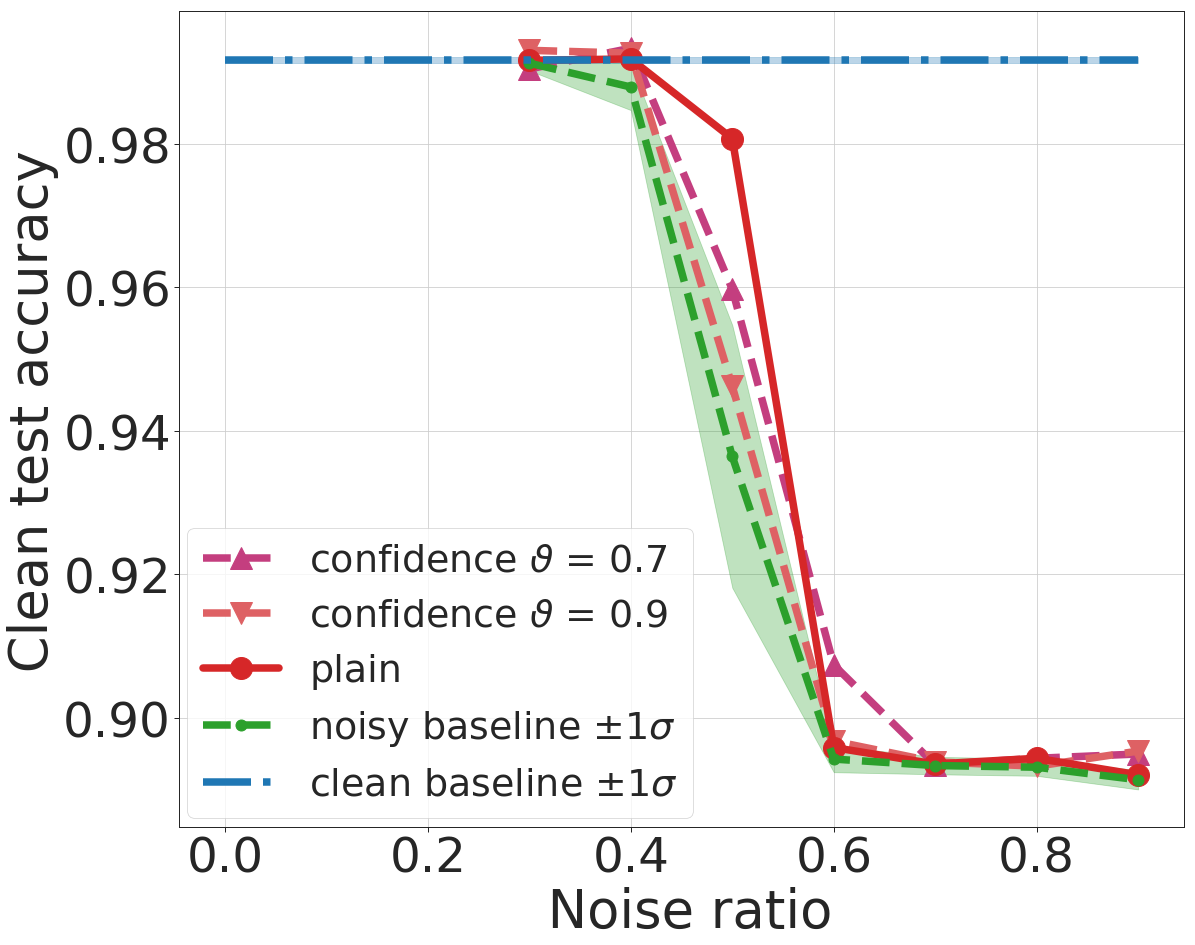}}
		\hfill
		\subfigure[MNIST-CNN on noisy MNIST data with bias error.]{\label{fig:mnist_bias_mnist}\includegraphics[width=0.24\textwidth]{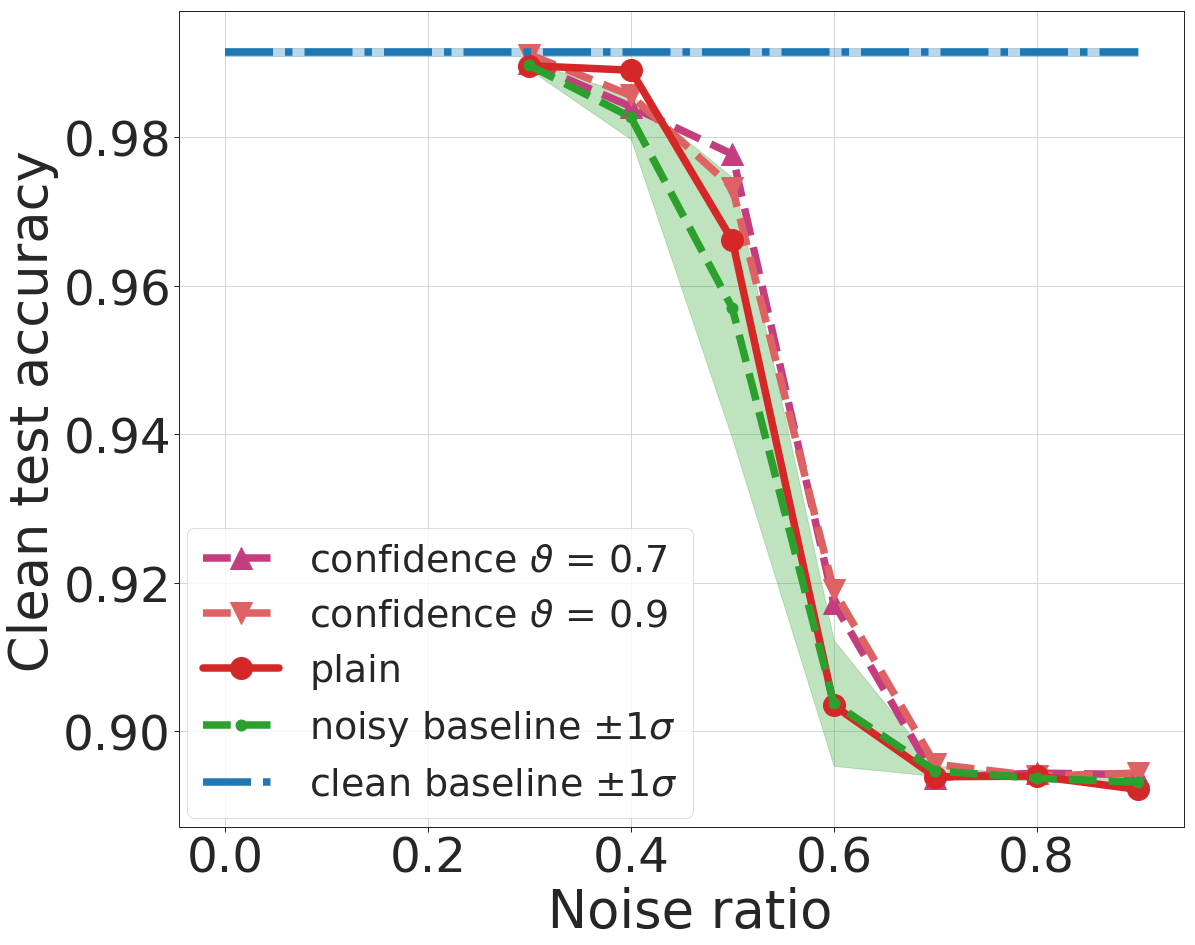}}
	\end{minipage}
	\caption{Different combinations of error type and network architecture and the performance of the ILI variants on noisy MNIST data. ILI is able to improve the accuracy in all tested settings.}
	\label{fig:mnist_combinations}
\end{figure*}

\begin{table*}[t]
	\caption{Relative improvement in accuracy from first to last ILI iteration of MNIST-CNN on MNIST data with random errors using plain ILI with initILI, 10 ILI iterations if not stated differently (frac. = fraction; conf. = confidence).}
	\label{tab:ilimnist}
	\vskip 0.15in
	\begin{center}
		\begin{small}
			\begin{sc}
				\begin{tabular}{cccccc}
					\rowcolor{lightgray!75}
					Noise frac. &  no filter [\%] &  no filter 50 ILI iterations [\%] &  conf. filter $\vartheta = 0.3$ [\%] & opILI [\%] & fpILI [\%] \\
					0.3 &         0.26 & - & - & \textbf{0.27} & 0.26 \\
					0.4 &         \textbf{0.69} & - & - & 0.60 & 0.38 \\
					0.5 &         0.95 & - & - & \textbf{1.67} & 0.37 \\
					0.6 &         2.77 & - & - & 3.32 & \textbf{4.05} \\
					0.7 &         7.56 & 8.02 & 5.64 & 11.5 & \textbf{14.2} \\
					0.8 &        34.35 & 38.62 & 25.6 & 42.1 & \textbf{51.2} \\
					0.9 &        48.24 & \textbf{69.79} & 21.5 & 22.4 & -5.87 \\
				\end{tabular}
			\end{sc}
		\end{small}
	\end{center}
	\vskip -0.1in
\end{table*}
\subsection{ILI for label correction}
We manually introduce two types of errors, (i) bias errors, meaning a class A is (if falsely labelled), labelled as B and never as C. (ii) random errors, all other classes but A are equally likely. For both types of errors and for a given noise fraction we randomly pick a subset of the original training labels provided with the datasets we use (all labels have equal probability of being picked), and for this subset we introduce the error as described above.

We apply the versions of ILI to MNIST \cite{lecun1998gradient} and CIFAR10 \cite{krizhevsky2009learning} with increasing amounts of erroneous labels. We use three different model architectures, two of which are defined in the Keras examples \cite{chollet2015keras}, which we call MNIST-CNN and CIFAR-CNN respectively and apply them both to noisy MNIST and noisy CIFAR10 training data.
Furthermore, we apply ILI to ResNet32 \cite{he2016deep} on noisy CIFAR10 data.
The noise fraction is the fraction of labels being changed as compared to the original, clean dataset. For the experiments we gradually increase the noise fraction, starting at $0.3$, in steps of $0.1$, up to $0.9$. If not otherwise noted, we use an early stopping approach, where we use the held-out clean validation set to terminate the optimisation iterations, if the validation accuracy is no longer increasing, with a maximum of ten iterations, if not stated differently. 

\subsubsection{Noisy MNIST data}
Figure \ref{fig:acc_noise_mnist} shows different versions of ILI using MNIST-CNN on noisy MNIST data, with random errors. All versions of ILI improve the accuracy over the noisy baseline in this setting. The best performing method in is ILI without partitioning, when run for 50 ILI iterations. All versions shown use refILI, as introduced in algorithm \ref{algo:opIliWithInitIli} for opILI.

Table \ref{tab:ilimnist} shows the relative improvement achieved by the ILI iterations, using MNIST-CNN on MNIST data with random errors.
Here, it should be noted that different versions of ILI use a different initialisation data set or ``seed" for the first training. Plain or filtering ILI with no partitioning
 uses a completely labelled training set, hence $N$ labels, but errors  in those labels are acceptable. On the other hand, using partitioning based ILI,
 e.g. opILI, reduces the amount of initial training labels used. If using $k$ partitions,
 and the same set, the seed will have a size of $N/k$, i.e. for opILI $N/2$.

Hence a comparison as in Figure \ref{fig:acc_noise_mnist} is not completely ``fair".
In terms of relative improvement, fragmentation based ILI (fpILI) outperforms all other methods,
except for a noise fraction of $0.9$, where
again the performance suffers from the smaller ``seed" training set. 

For plainILI with filter, with confidence as a metric, we observe that the performance is highly threshold dependent. Thresholds $\vartheta = 0.5, 0.7, 0.9$ lead to no improvement at all,
while $\vartheta = 0.3$ improves the accuracy over the optimisation iterations. However, if compared to ILI without filter, the performance of ILI filter is inferior.
To find out which threshold setting achieves the best performance, the expectation should be taken into account that in the later ILI iterations, the performance
of the predicted labels is getting higher, so a larger fraction of samples should be selected in comparison to earlier ILI iterations.
On the other hand, the confidence itself is expected to increase anyway together with the model accuracy during the course of the ILI process. However, this is complicated by the fact that the confidence is likely to depend additionally on the number of training iterations performed
per ILI iteration. Therefore, an effective choice of the threshold, which may vary throughout the process,
has the potential to further increase the overall performance of the ILI approach but remains part of future work.
Figure \ref{fig:mnist_combinations} shows the results of different ILI variants applied to noisy MNIST data, with random error in Figure \ref{fig:mnist_random_cifar}, using CIFAR-CNN and Figure \ref{fig:mnist_random_mnist}, using MNIST-CNN. The results for bias errors are shown in Figure \ref{fig:mnist_bias_cifar}, for CIFAR-CNN and Figure \ref{fig:mnist_bias_mnist}, for MNIST-CNN. In all settings ILI is able to improve the results.

\subsubsection{Noisy CIFAR10 data}\label{subsubsec:cifar}
\begin{figure*}[htpb]
	\centering
	\begin{minipage}{0.99\linewidth}
		\centering
		\subfigure[CIFAR-CNN on noisy CIFAR10 data with random error]{\label{fig:cifar_random_cifar}\includegraphics[width=0.24\textwidth]{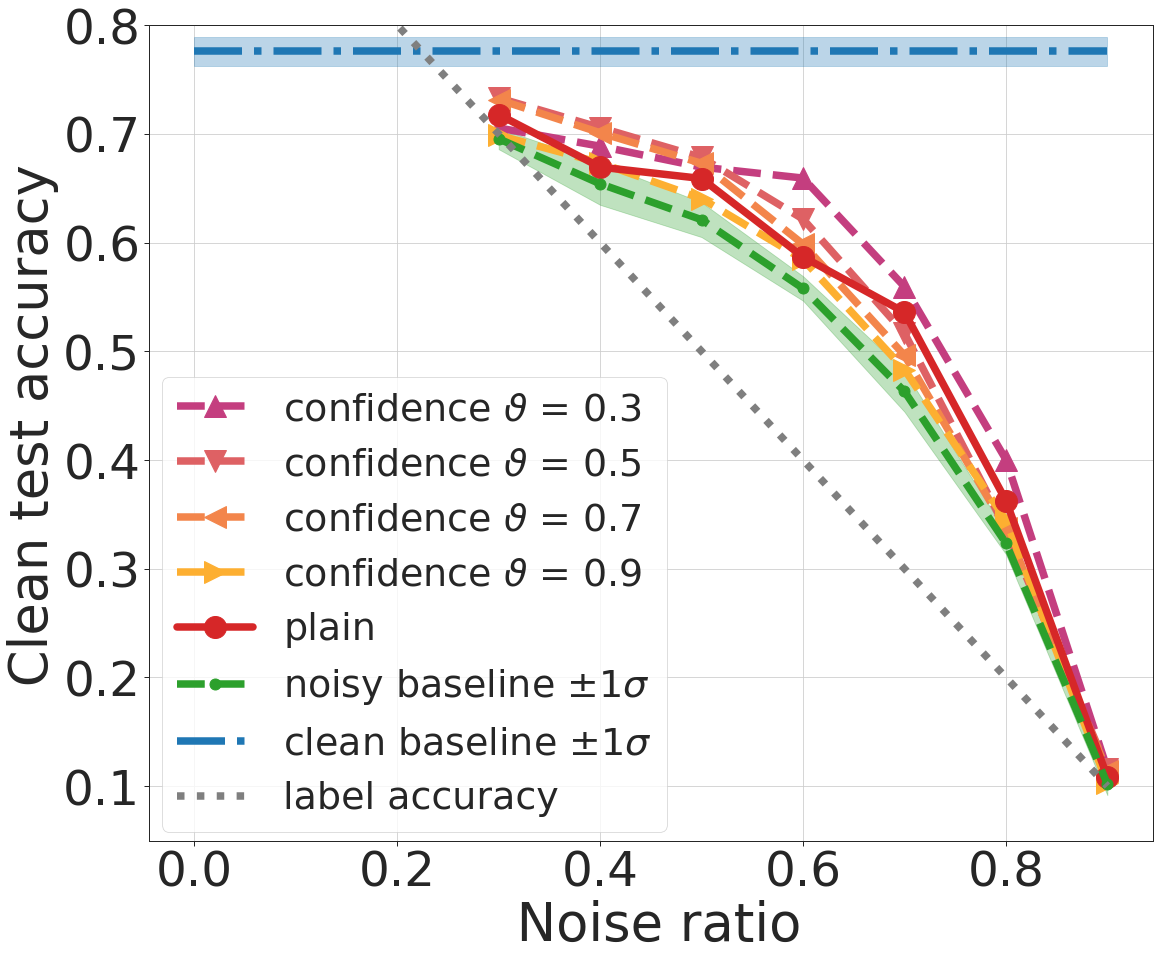}}
		\hfill
		\subfigure[MNIST-CNN on noisy CIFAR10 data with random error. ]{\label{fig:cifar_random_mnist}\includegraphics[width=0.24\textwidth]{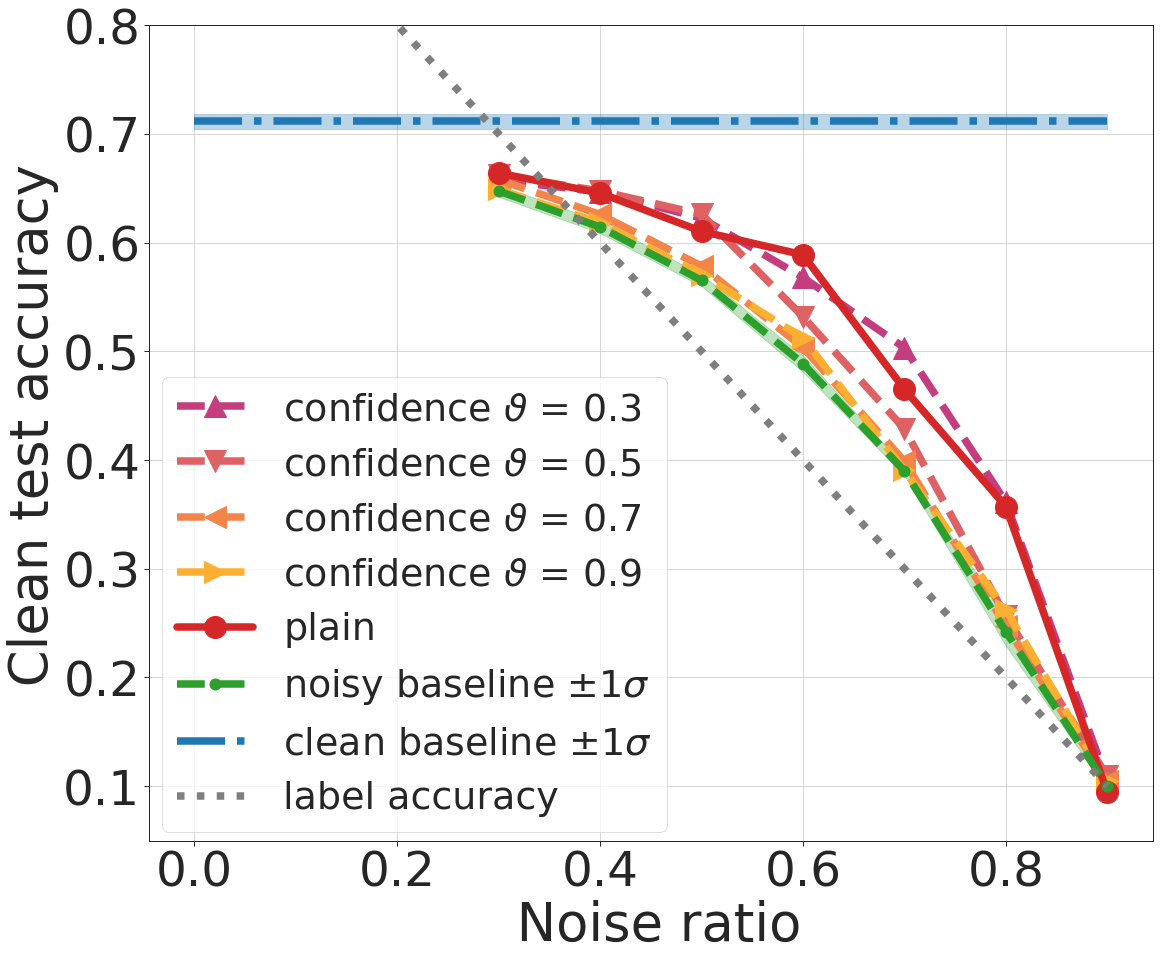}}
		\hfill 
		\subfigure[CIFAR-CNN on noisy CIFAR10 data with bias error. ]{\label{fig:cifar_bias_cifar}\includegraphics[width=0.24\textwidth]{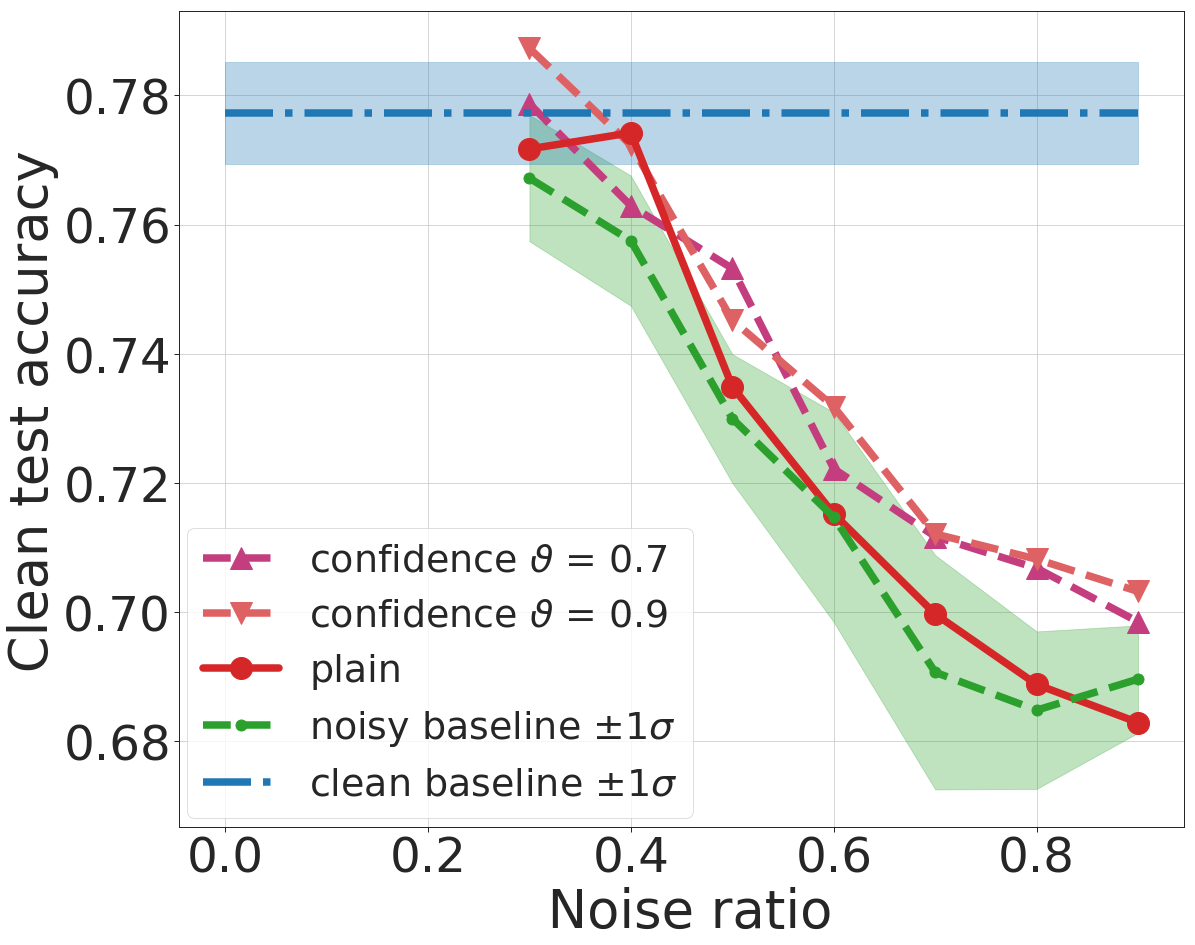}}
		\hfill
		\subfigure[ResNet on noisy CIFAR10 data with bias error. ]{\label{fig:cifar_bias_resnet}\includegraphics[width=0.24\textwidth]{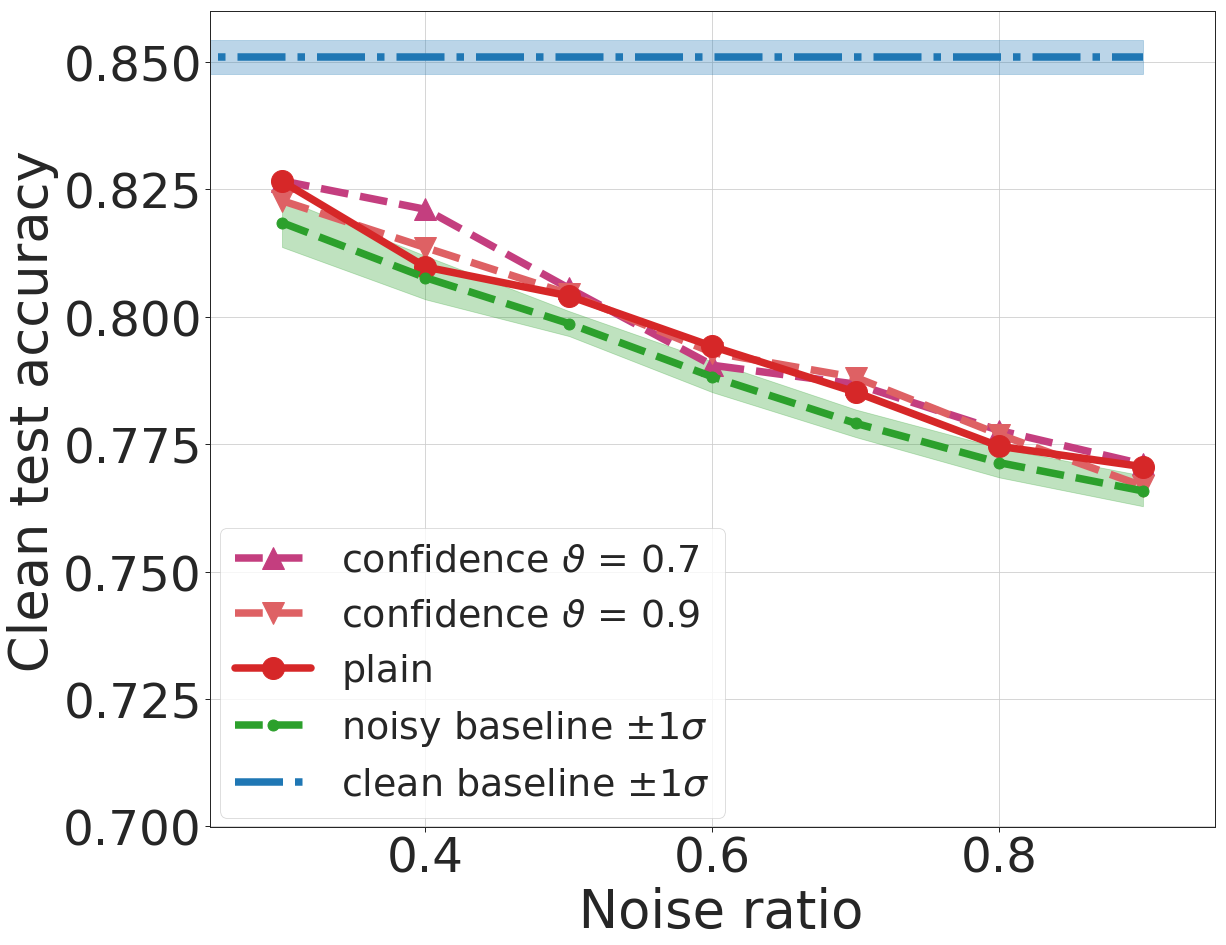}}
	\end{minipage}
	\caption{Different combinations of error type and network architecture and the performance of the ILI variants on noisy CIFAR10 data. ILI is able to improve the accuracy in all tested settings.}
	\label{fig:cifar}
\end{figure*}

For CIFAR10, the accuracies achievable by the chosen models and hyperparameters are considerably lower than on MNIST. However, from Figure \ref{fig:cifar} we can see that the different ILI variants presented improve the accuracy for a given noise fraction. This holds true for both bias and random errors. Figures \ref{fig:cifar_random_cifar} (CIFAR-CNN) and \ref{fig:cifar_random_mnist} (MNIST-CNN) show the results for random errors, while figures \ref{fig:cifar_bias_cifar} (CIFAR-CNN) and \ref{fig:cifar_bias_resnet} (ResNet32) show the results for bias errors. \\
As a reference, we compare our approach using a ResNet32 \cite{he2016deep} to the ``learning to reweight" technique \cite{ren2018learning} using the very same network architecture. As ILI can be combined with the ``learning to reweight" approach, we do not need or expect to outperform this method. We use the code provided by the authors\footnote{\url{https://github.com/uber-research/learning-to-reweight-examples}} to reproduce the results for random errors (called \textit{uniform flip} by the authors). The results can be seen in Figure \ref{fig:reweight} and \ref{fig:reweight_noaug}. If using data augmentation, our approach outperforms ``learning to reweight" for the noise regime from $0.3$ to $0.7$. For a noise fraction of $0.8$ and higher, our approach fails to generalize as well and the performance is below that of ``learning to reweight", as shown in Figure \ref{fig:reweight}. If data augmentation is deactivated, the results for ``learning to reweight" are slightly better than ours, cf. Figure \ref{fig:reweight_noaug}. However, without data augmentation the used ResNet32 as a network with much higher capacity in comparison to MNIST-CNN or CIFAR-CNN easily overfits the noisy training data. Hence it fails to generalise on clean validation data. The performance on the clean validation data is inferior to the label accuracy, hence our iterative approach fails for most noise fractions as expected. Note that ``learning to reweight" also leads to a validation accuracy inferior to the labelling quality. For a noise fraction of $0.7$ and $0.8$, our confidence based filtering ILI with thresholds $\vartheta = 0.9$ outperforms ``learning to reweight" and increases the accuracy over the training label accuracy, even without data augmentation. 

\subsection{ILI for semi-supervised learning}

\begin{figure*}[tpb]
	\centering
	\begin{minipage}{0.99\linewidth}
		\centering
		\subfigure[ILI vs. ``learning to reweight" \cite{ren2018learning}. Our method outperforms ``learning to reweight" for noise fractions $<0.8$. ]{\label{fig:reweight}\includegraphics[width=0.24\textwidth]{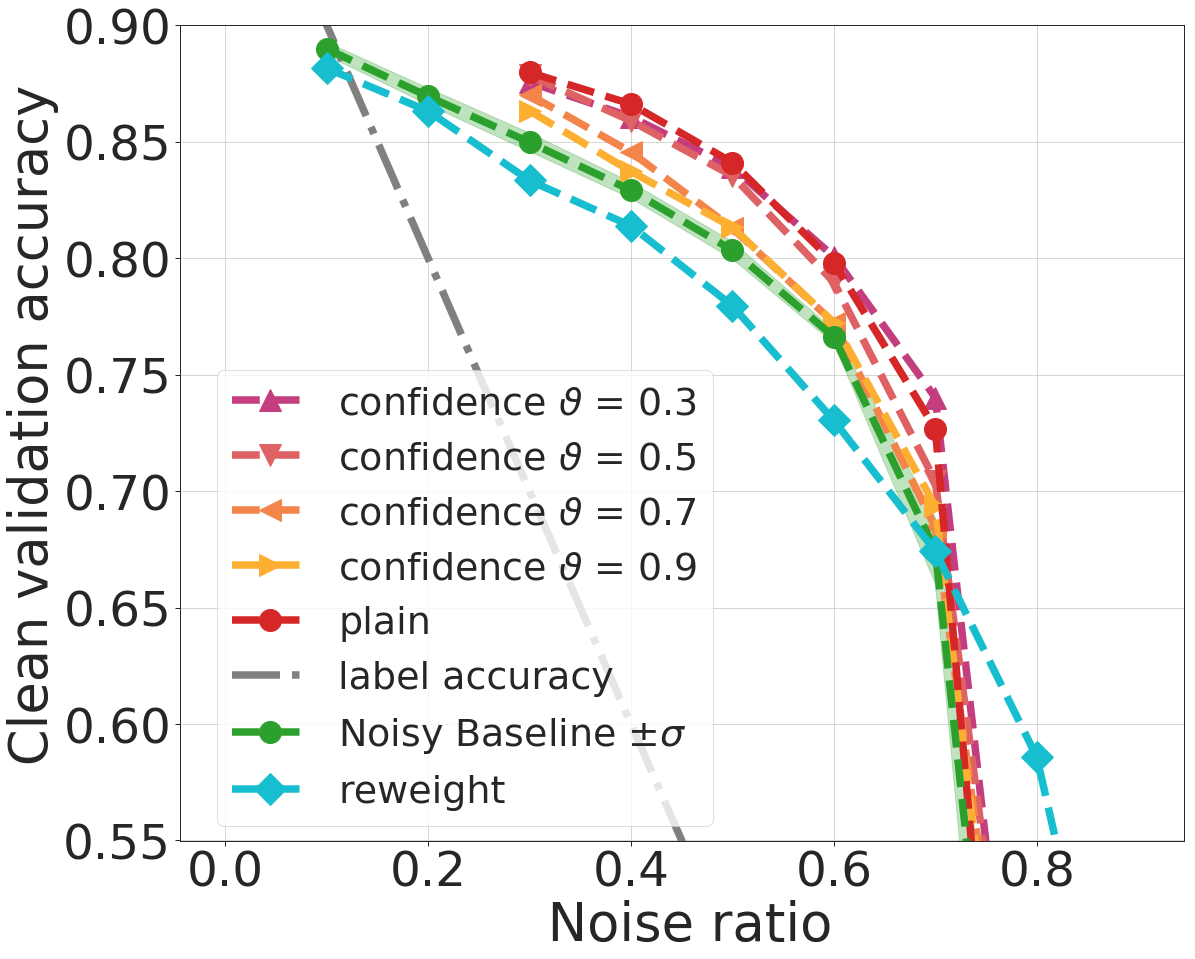}}
		\hfill
		\subfigure[ILI vs. ``learning to reweight". Without data augmentation both methods fail to improve the accuracy significantly. ]{\label{fig:reweight_noaug}\includegraphics[width=0.24\textwidth]{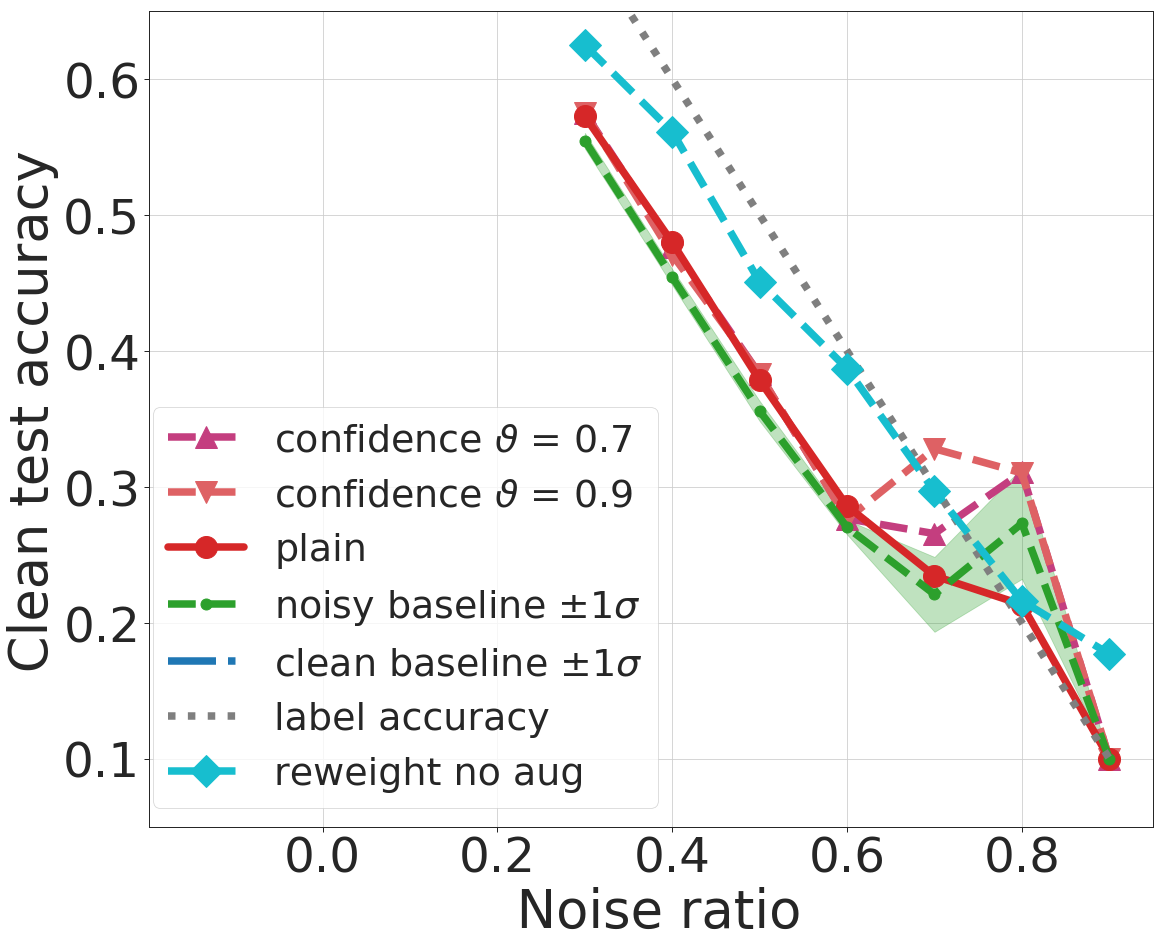}}
		\hfill
		\centering
		\subfigure[ILI for MNIST-CNN on noisy MNIST data (random error), SSL, with final training (FT). ]{\label{fig:acc_noise_mnist_final}\includegraphics[width=0.24\textwidth]{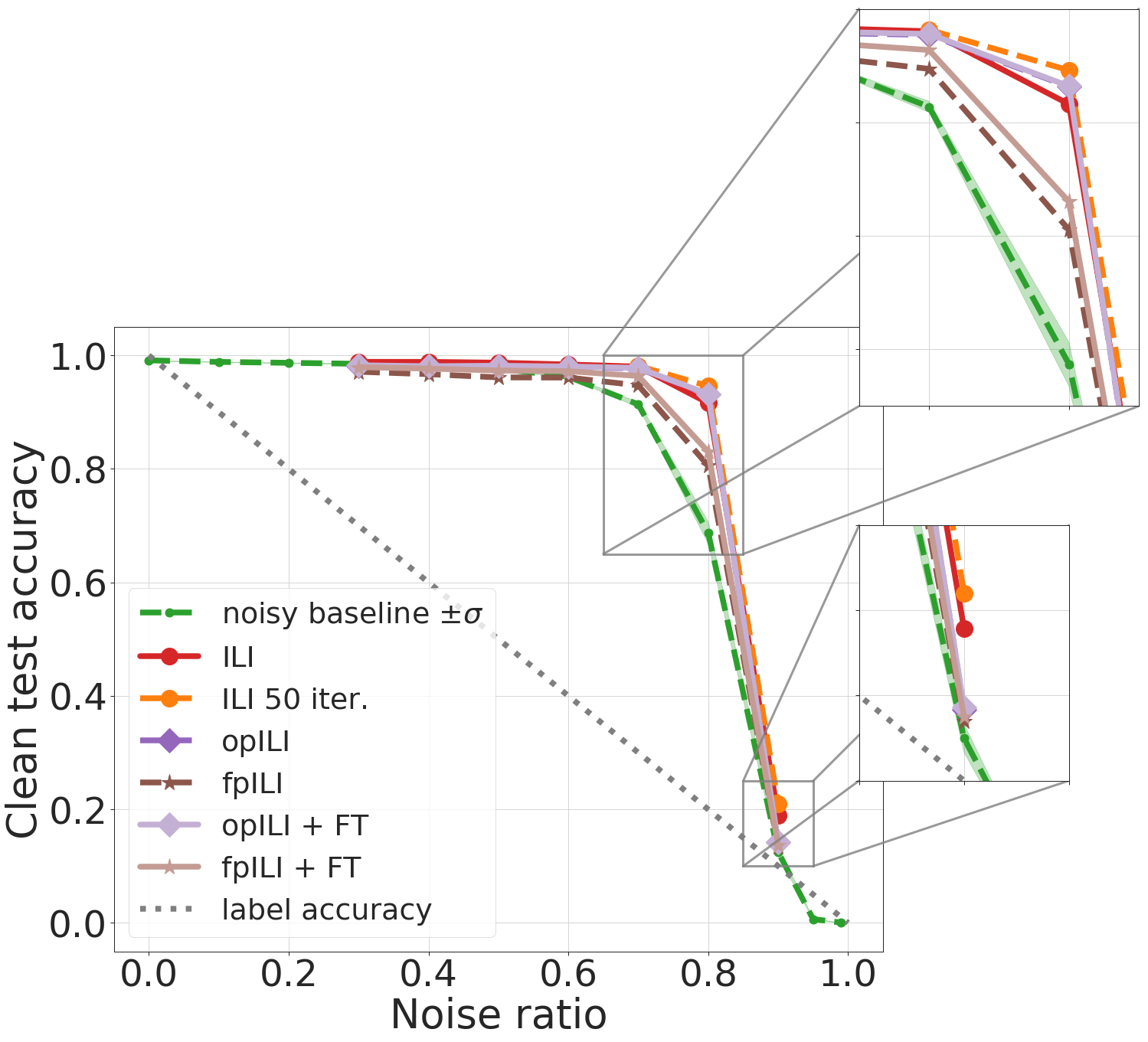}}
		\hfill
		\subfigure[SSL, comparison using half the dataset only, opILI leverages the unlabelled data and achieves superior performance.]{\label{fig:opili_half}\includegraphics[width=0.24\textwidth]{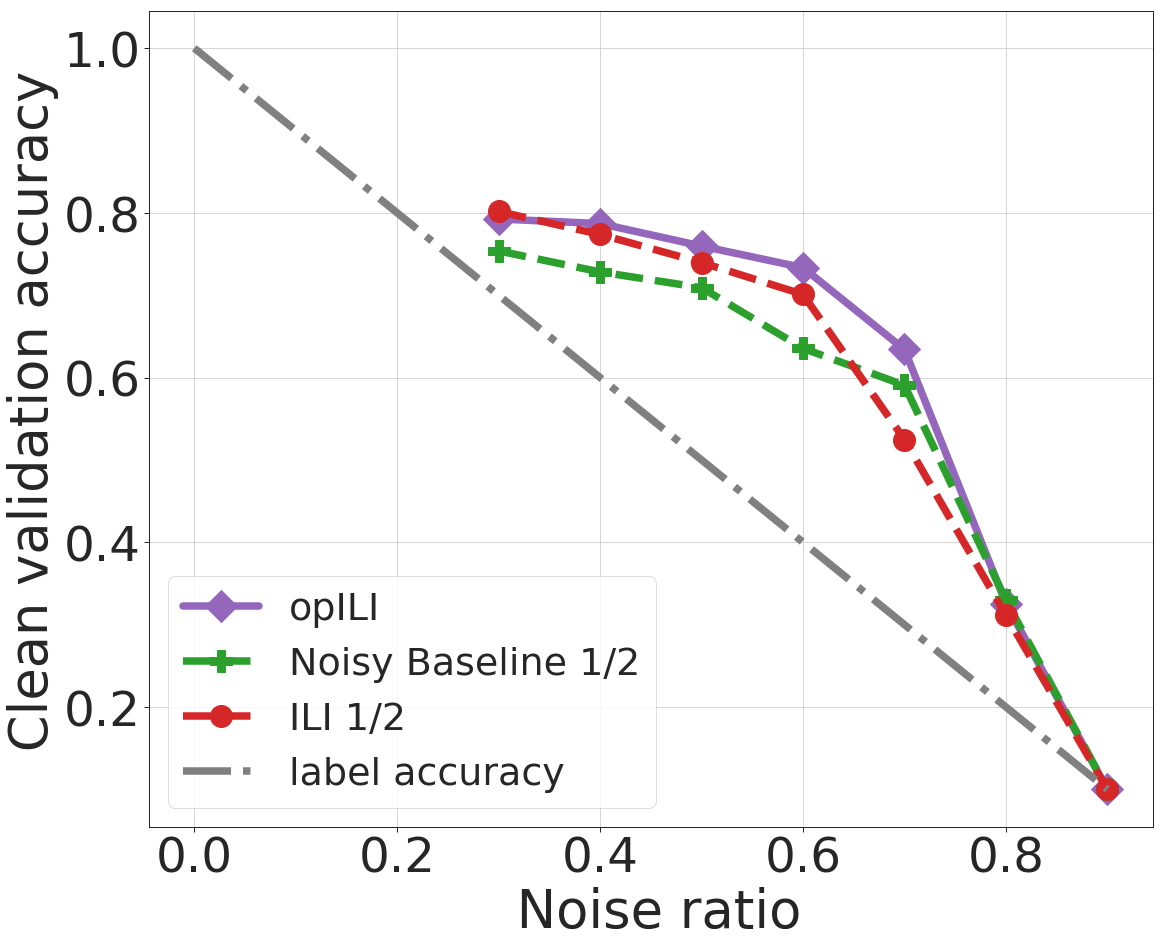}}
	\end{minipage}
	\caption{Comparison of different versions of our ILI algorithm on CIFAR10 data, using a ResNet32, with erroneous labels (randomly distributed) vs. ``learning to reweight" (left). And ILI variants in a SSL setting (right).}
	\label{fig:mnist_combinations}
\end{figure*}

Partitioning based ILI (pILI) as introduced in Section \ref{subsec:algo:pili} is suitable for SSL, as it only requires labels for the first partition, while gradually improving pseudo-labels one the other partition(s). A final training using all the subsets with their best predicted labels can be performed in order to gain more accuracy. The final training is conducted on the set
\[X\lt{train, A} \cup \left(\bigcup_i X\lt{train, B$_i$}\right).\]
(note: This includes the opILI case where $\{B_i\} = B$).\\
The resulting performance plot for the MNIST-CNN on the MNIST dataset with random errors is shown in Figure \ref{fig:acc_noise_mnist_final}. The results are compared in Table \ref{tab:ilimnistfinal} by their percentage increase in accuracy from the first to the last ILI iteration. A final training on all available data can lead to minor performance improvements. All variants increase the validation accuracy, however fpILI is inferior, due to the smaller intilization set, as discussed in sections \ref{subsec:algo:pili} and \ref{subsec:ili_real}. 

\begin{table*}[t]
	\caption{Relative improvement in accuracy from first to last ILI iteration of MNIST-CNN on MNIST data with random errors using plain ILI with initILI, 10 ILI iterations if not stated differently (frac. = fraction; conf. = confidence; FT = final training).}
	\label{tab:ilimnistfinal}
	\vskip 0.15in
	\begin{center}
		\begin{small}
			\begin{sc}
				\begin{tabular}{cccccccc}
					\rowcolor{lightgray!75}
					Noise &  No Filter [\%] &  No Filter &  Conf. Filter & opILI [\%] & opILI+FT [\%] & fpILI [\%] & fpILI+FT [\%] \\
					\rowcolor{lightgray!75}
					frac. &   &  50 ILI Iter. [\%] &  $\vartheta=0.3$ [\%] &  &  &  & \\
					0.3 &         0.26 & - & - & 0.27 & 0.34 & 0.26 & \textbf{1.1} \\
					0.4 &         0.69 & - & - & 0.60 & 0.72 & 0.38 & \textbf{1.4} \\
					0.5 &         0.95 & - & - & 1.67 & \textbf{1.75} & 0.37 & 1.64 \\
					0.6 &         2.77 & - & - & 3.32 & 3.44 & 4.05 & \textbf{5.26} \\
					0.7 &         7.56 & 8.02 & 5.64 & 11.5 & 11.6 & 14.2 & \textbf{16.2} \\
					0.8 &        34.35 & 38.62 & 25.6 & 42.1 & 42.1 & 51.2 & \textbf{55.8} \\
					0.9 &        48.24 & \textbf{69.79} & 30.9 & 21.5 & 22.4 & -5.87 & -3.98 \\
				\end{tabular}
			\end{sc}
		\end{small}
	\end{center}
	\vskip -0.1in
\end{table*}



To enable a fair comparison between opILI and ILI, Figure \ref{fig:opili_half} shows the resulting performance of a ResNet32 trained on noisy CIFAR10 data with random error and a reduced training dataset size. For the noisy baseline and the ILI variant without partitioning, only half the dataset is used. For opILI, only half the dataset is labeled, the other half is used as unlabelled set to leverage the potential of opILI as SSL technique. Under these constrains opILI outperforms ILI without partitioning.

\subsection{ILI in real-world applications}\label{subsec:ili_real}
In our experiments, even though plain ILI (with or without filtering) is the most basic technique, the best results are achieved with this variant for noisy MNIST data.
This demonstrates the feasibility of the ILI approach in principle, as in the case of this simple classification task,
no further measures are necessary in order to exploit iteratively corrected labels.
When considering the results of the partitioning based variants, the experimental design
should be taken into account. The smaller initialisation set appears to
outweigh the advantage of predictions on unseen data in this example, as 
these more sophisticated variants do not show an additional improvement over plainILI.
In our experimental setting we utilise a given dataset to investigate the performance on this dataset for which we know the correct labels.
In practice it would rather be typical to have a small labelled set and larger amounts of unlabelled data.
Using fpILI or opILI does not require initial labels for any other but the first subset.
Hence, the strength of these partitioning based variants will become apparent in applications where larger amounts of unlabelled data can be collected easily.

\section{Conclusions}

We introduced the ILI approach and its variants, which iteratively improve erroneous labels. Our approach is inspired by recent works on semi-supervised learning \cite{yalniz2019billion}, self-training \cite{xie2020self} and iterative cross learning \cite{yuan2018iterative}. 
Being architecture agnostic, ILI can be applied to various deep neural networks. We illustrate the usage of ILI with three different architectures. Furthermore, ILI can be combined with approaches for regularisation
to improve generalisation, in order to take advantage of their full potential. If provided with a clean, well-labelled reference set,
ILI can leverage the information contained therein, given substantial additional amounts of inexpensive, unlabelled data so that ILI can
integrate those samples and generate labels. We successfully apply ILI in an SSL setting and show how using a confidence filter can benefit self-training.

By introducing controlled amounts of error to two well-known data sets and applying ILI to different network architectures, we show its applicability.
More complex filtering schemes based on various uncertainty measures have the potential to improve the results further.
Training schemes suitable for settings where erroneous labels are to be expected can be integrated with ILI to raise the reachable accuracy. 
Increasing the fraction of erroneous labels from $0.3$ in steps of $0.1$ to $0.9$, we find that applying ILI consistently improves the accuracy on the clean test set
in comparison to performing a regular training with the erroneous labels and without iterations.
The only exceptions are the cases when the fraction of erroneous labels is very large, e.g. a noise fraction of $0.9$ for MNIST, or when the initial training fails to generalise well enough to lead to a performance above the training label accuracy. 
We compare ILI to ``learning to reweight", another technique dealing with erroneous labels (which could be combined with ILI) and find that regularisation is a key ingredient to any approach dealing with erroneous labels. We show so, by applying data augmentation. If data augmentation is applied, our approach outperforms ``learning to reweight", which is as state-of-the-art method. We hypothesize that any approach successfully dealing with a noisy label setting depends on the ability to generalise well from the uniformly distributed errors. However if such a generalisation fails in the first place i.e. the noisy training data is overfitted by a high capacity neural network, regularisation is essential. \\
Therefore, we expect that ILI can be improved additionally by further investigations into regularisation in settings with erroneous labels, which is part of future work. Furthermore, our approach can be combined with uncertainty measures more sophisticated than the confidence derived simply from neural network output activations. Also combining other recent techniques to learn from erroneous labels with ILI and showing the applicability to a broader range of network architectures, as well as larger datasets like ImageNet \cite{deng2009imagenet} will raise interesting questions.

\bibliography{bibliography}
\bibliographystyle{IEEEtran}

\newpage

\appendices

\section{The effect of erroneous labels}

\begin{figure*}[tpb]
	\centering
	\begin{minipage}{0.99\linewidth}
		\centering
		\subfigure[MNIST-CNN on noisy MNIST data, with random error. ]{\label{app:fig:multi_noise_mnist}\includegraphics[width=0.32\textwidth]{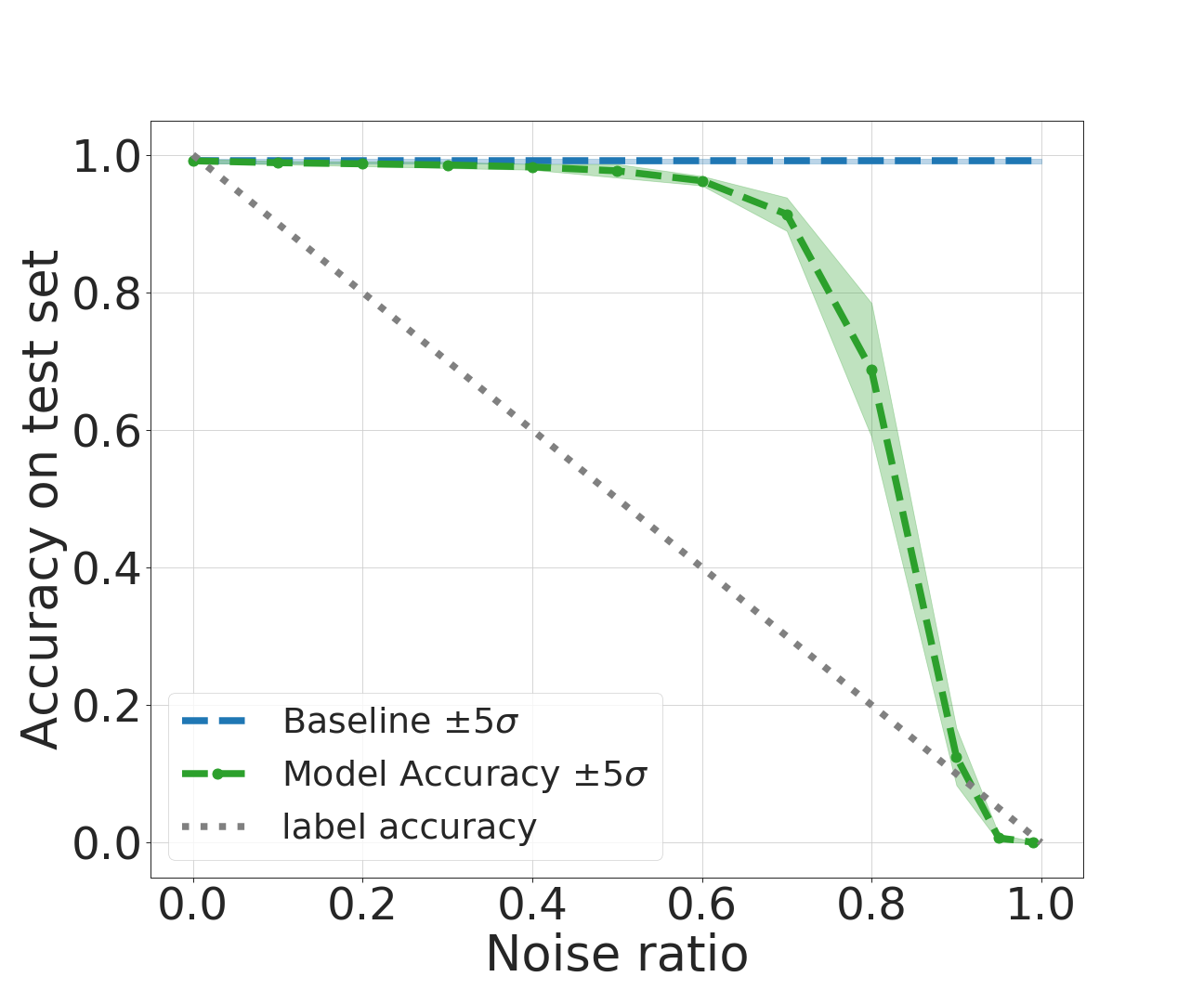}}
		\hfill
		\subfigure[CIFAR-CNN on noisy CIFAR data, with random error. ]{\label{app:fig:multi_noise_cifar}\includegraphics[width=0.32\textwidth]{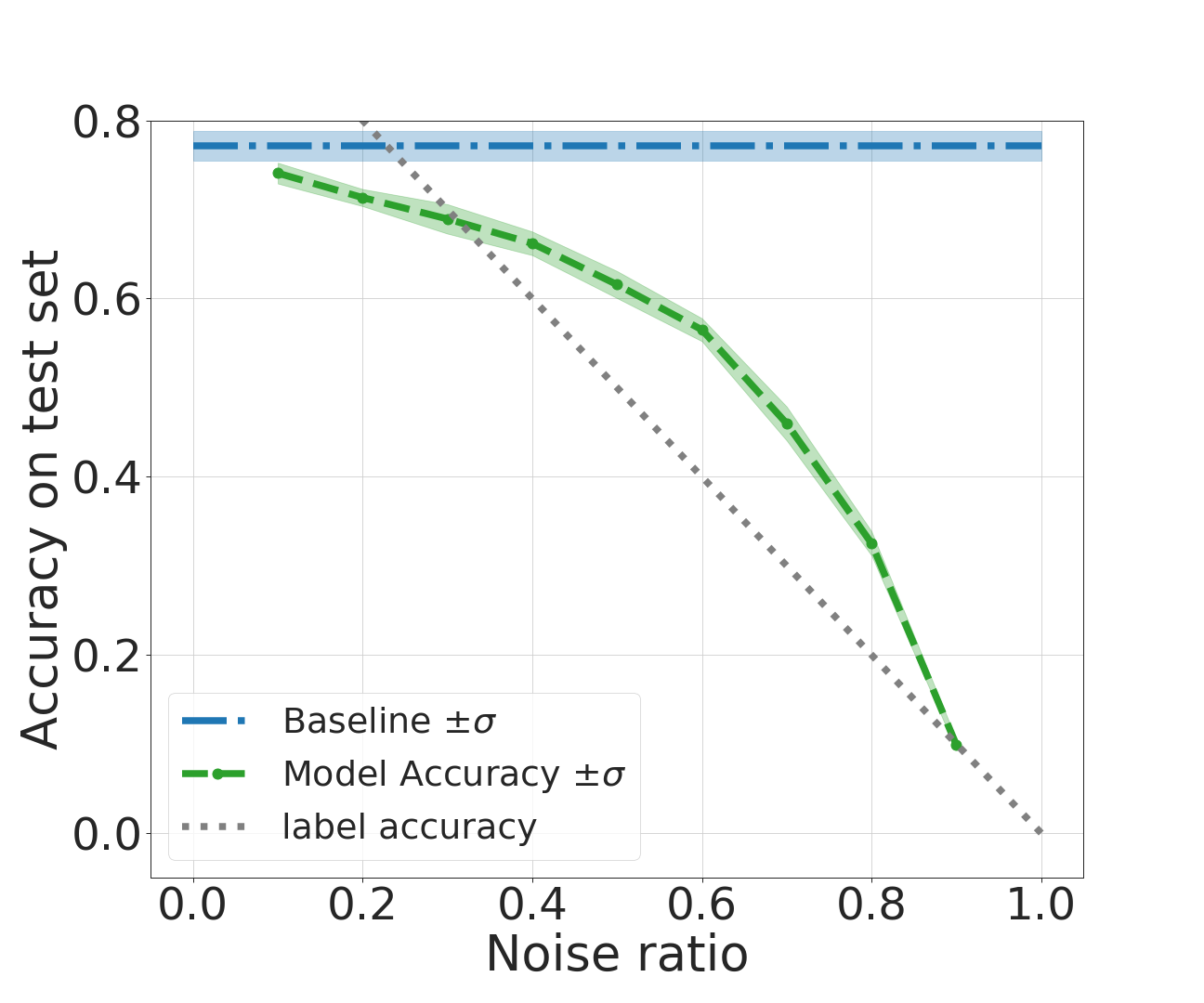}}
		\hfill
		\subfigure[MNIST-CNN on noisy MNIST data, with bias error. ]{\label{app:fig:multi_noise_mnist_bias}\includegraphics[width=0.32\textwidth]{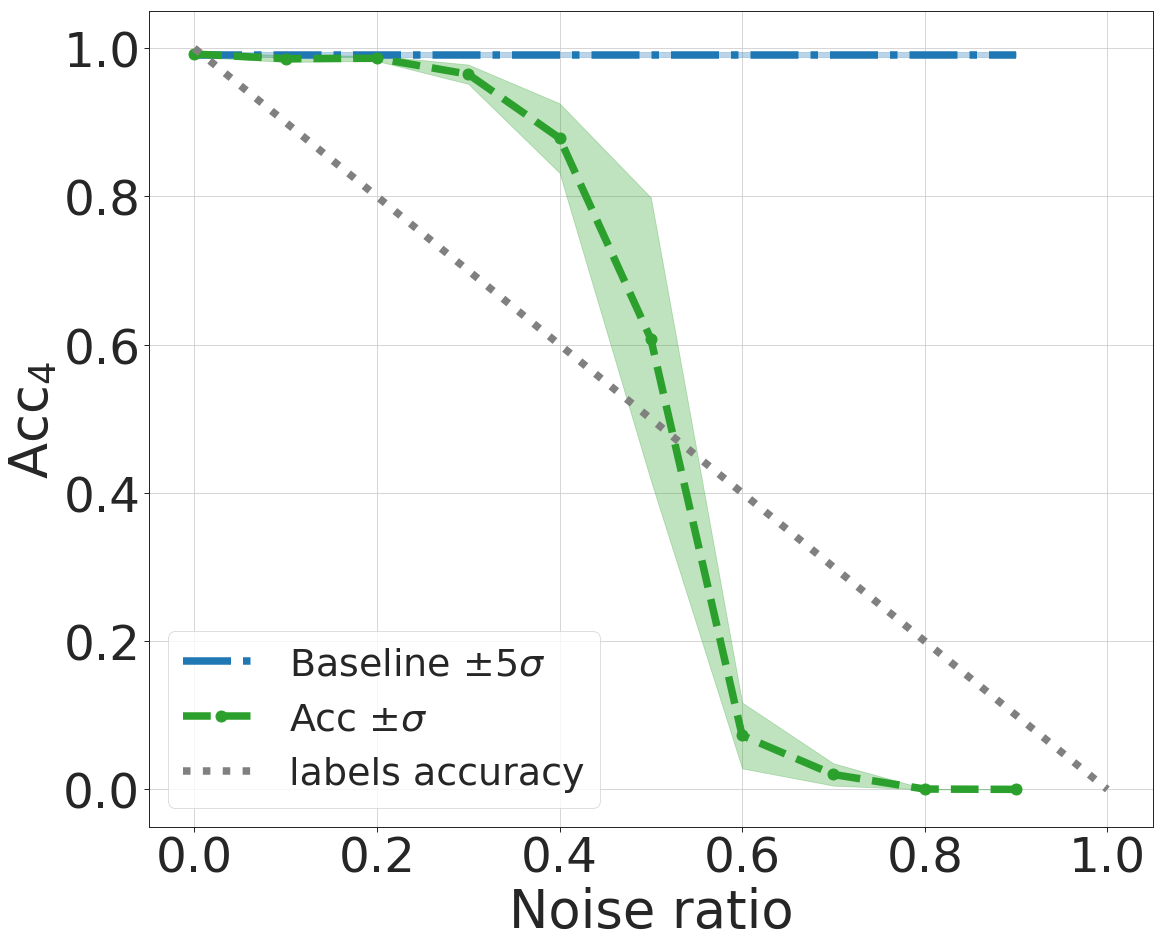}}
	\end{minipage}
	\caption{The effect of different fractions and types of error (random \& bias) on the performance of a network trained with these erroneous labels.} 
\end{figure*}

In this section we give a detailed overview on the effects of the amount of error in the training material on the resulting accuracies on a clean test set, this complements section II in the main document. Figure \ref{app:fig:multi_noise_mnist} shows the resulting performance if MNIST-CNN is trained on noisy MNIST data, with random error. Only above $0.9$ the MNIST-CNN is not able to generalise well, and does not achieve a resulting accuracy larger than the accuracy of the training labels. The empirical variation in the resulting performance is largest for $0.7$ and $0.8$. Figure \ref{app:fig:multi_noise_cifar} shows the resulting performance for CIFAR-CNN, trained on noisy CIFAR10 data with random error. Only for noise fractions from $0.4$ to $0.8$ the resulting performance is higher than the accuracy of the training labels, hence only in this regime we can expect our ILI approach to be successful. In Figure \ref{app:fig:multi_noise_mnist_bias} the results on noisy MNIST data with bias error, using a MNIST-CNN are shown. The accuracy $\text{Acc}\lt{4}$ is the accuracy on samples labelled with class ``4" from the test set\footnote{The bias for this experiment was switching the given fraction of samples from class ``4" to ``7".} For the interval $[0.0, 0.5]$ the network is able to achieve an accuracy above the accuracy of the training labels, despite the bias error. However, above a noise fraction of $0.5$, the majority of samples with the original class ``4" are falsely labelled with ``7", hence the performance drops to almost zero\footnote{Also in this case the performance on the ``4"s from the test set, if using the complete test set, the accuracy would drop to $\approx 0.9$ for ten classes.} above this point.

\section{Detailed results over the optimisation iterations}
To further illustrate the results summarized in the main document in the Figures 1, 3-5 and in section V, we show the resulting test accuracy over the optimisation iterations.

\begin{figure}[tbp]
	\centering
	\centering
	\includegraphics[width=0.99\linewidth, draft=false]{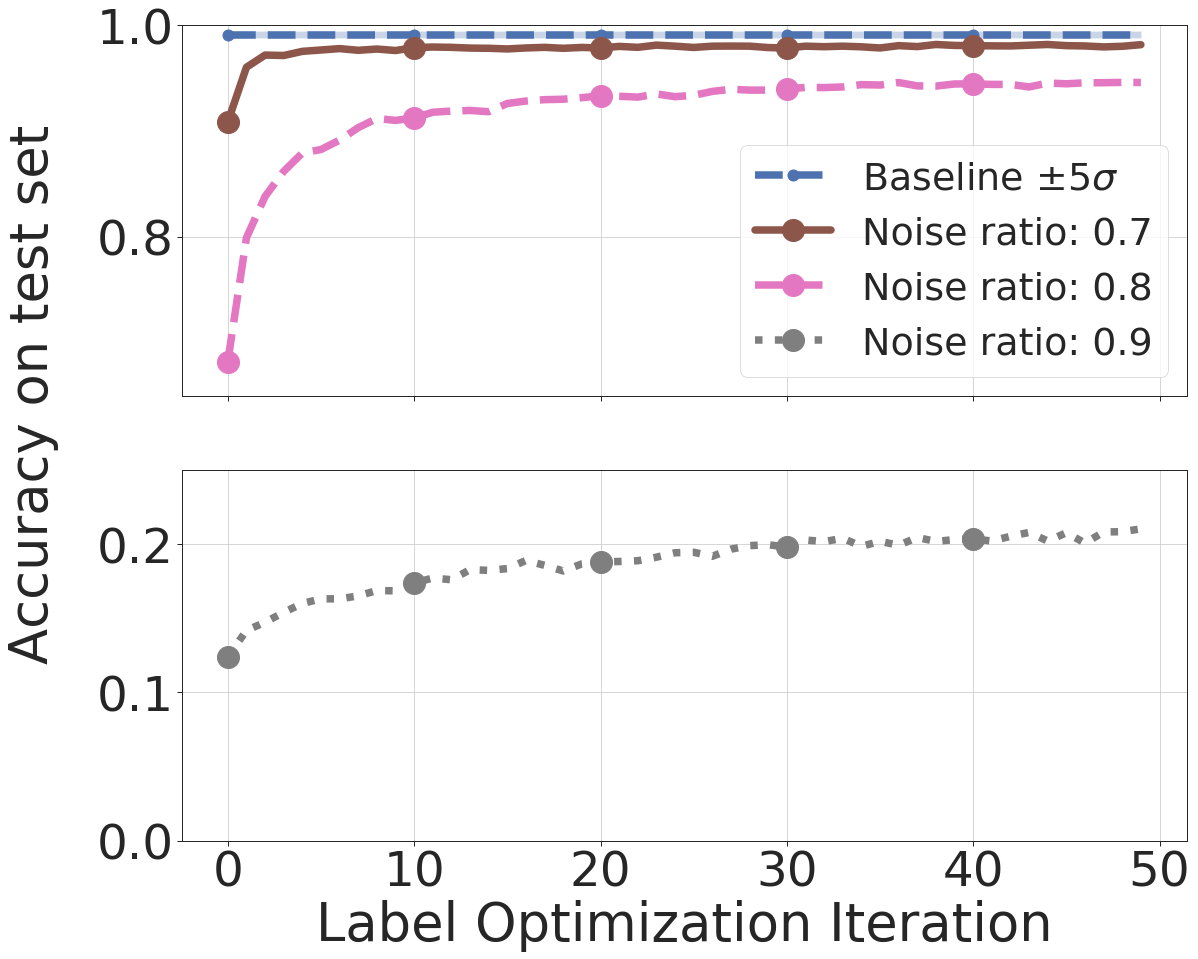}
	\caption{ILI Plain 50 optimisation iterations on MNIST.  
		For reference, the maximum reachable accuracy for the given model is shown (dashed blue line). The test accuracy is increasing over the iterations. 
		This holds true for all investigated levels of erroneous initialisation labels, even for the highest level considered, i.e. a fraction of $0.9$ faulty
		labels.}\label{fig:iliplain_50iter}
	\vspace{-2.5mm}
\end{figure}

Figure \ref{fig:iliplain_50iter} shows the resulting test accuracy over 50 optimisation iterations of ILI without filter, using MNIST-CNN on MNIST data with random error and initILI. The maximum accuracy with the $\pm \sigma$ interval is shown as well. This is the maximum accuracy we can reach with the given model and dataset (without additional hyperparameter tuning). Noise ratios are shown from $0.7$ to $0.9$. For all noise ratios, the test accuracy is increasing over the optimisation iterations, demonstrating ILI's effectiveness over a wide range of noise levels. In the intermediate noise regime, at noise ratios of $0.8$ and $0.7$, the largest increase in test accuracy is achieved. For better visibility, smaller noise fractions ($<0.7$) are not shown. At all noise fractions, the largest gain in performance is achieved within the first few iterations of ILI.
Figure \ref{fig:iliplain_50iter} shows that for a noise fraction of $0.8$, performance saturates between $30$ and $40$ iterations and
for $0.9$, saturation is reached after $40$ iterations. For noisy MNIST data, using MNIST-CNN the test accuracy is increasing over the optimisation iterations, for all noise fractions investigated. This is a reasonable outcome considering the dataset size, as a fraction of $0.9$ erroneous labels, without any bias,
for the MNIST dataset means that there are still $6000$ correctly labelled samples in the training set.
From these, the network is able to generalise and correct some of the other errors present.
The increase in performance shows an asymptotic behaviour with the biggest gain occuring after the first iteration.

\begin{figure}[tbp]
	\centering
	\centering
	\includegraphics[width=0.99\linewidth, draft=false]{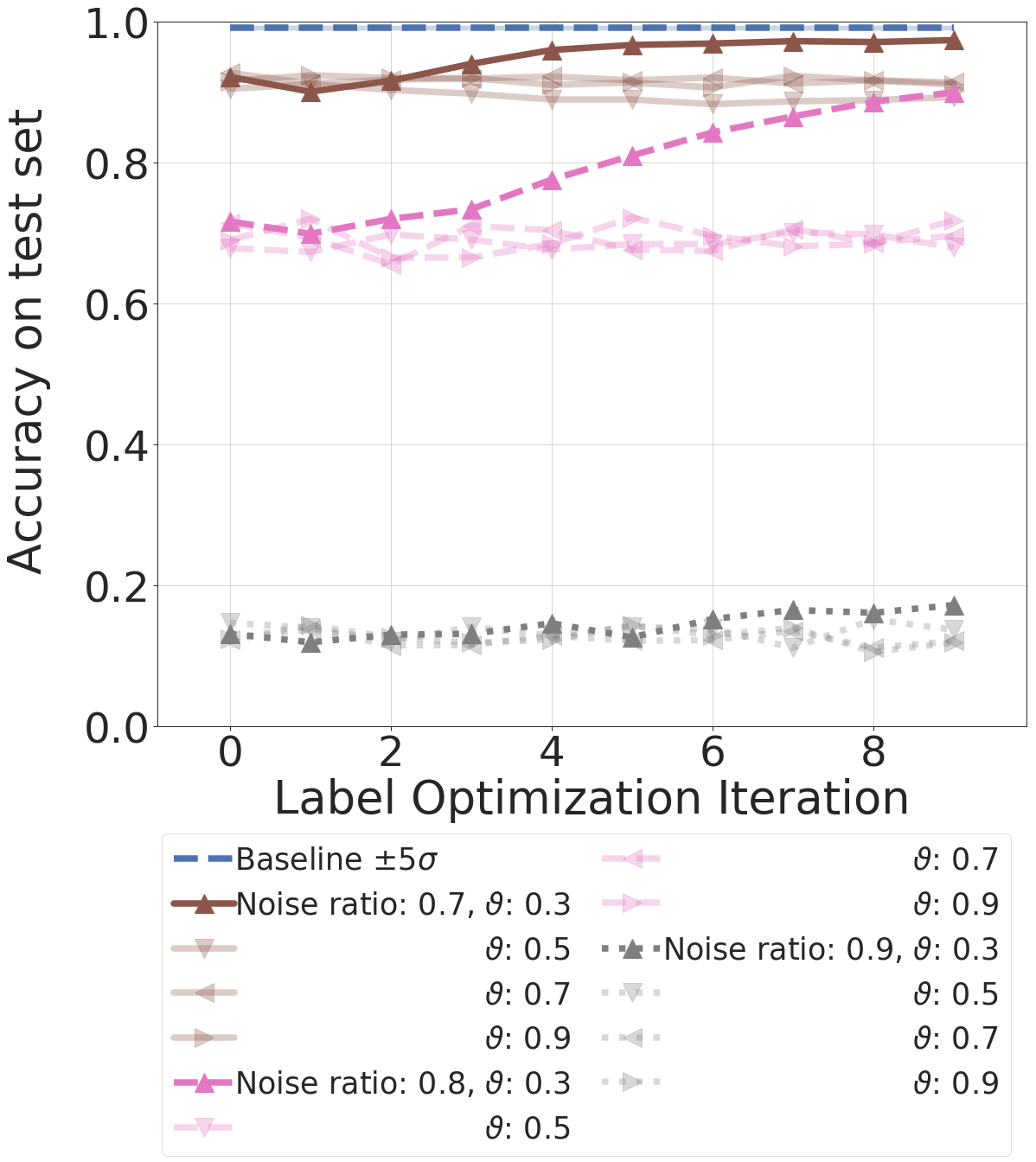}
	\caption{ILI with confidence filter and 10 optimisation iterations using MNIST-CNN on noisy MNIST data with random error. Confidence based filtering is highly threshold dependent. In the experiments on noisy MNIST data with MNIST-CNN and random error, only $\vartheta = 0.3$ leads to an improvement.}\label{fig:iliplain_10iter_confidence}
	\vspace{-2.5mm}
\end{figure}

Figure \ref{fig:iliplain_10iter_confidence} shows detailed results for all tested confidence thresholds ($0.3$, $0.5$, $0.7$ and $0.9$). The results on noisy MNIST data with randomly distributed errors using MNIST-CNN lead to the conclusion, that $\vartheta = 0.3$ is to be preferred, as it leads to the best results. The threshold $\vartheta$ is a hyperparameter, for which to tune a validation set (possibly noisy) is necessary.

\section{Overfitting ResNet}
\begin{figure}[tbp]
	\centering
	\centering
	\includegraphics[width=0.99\linewidth]{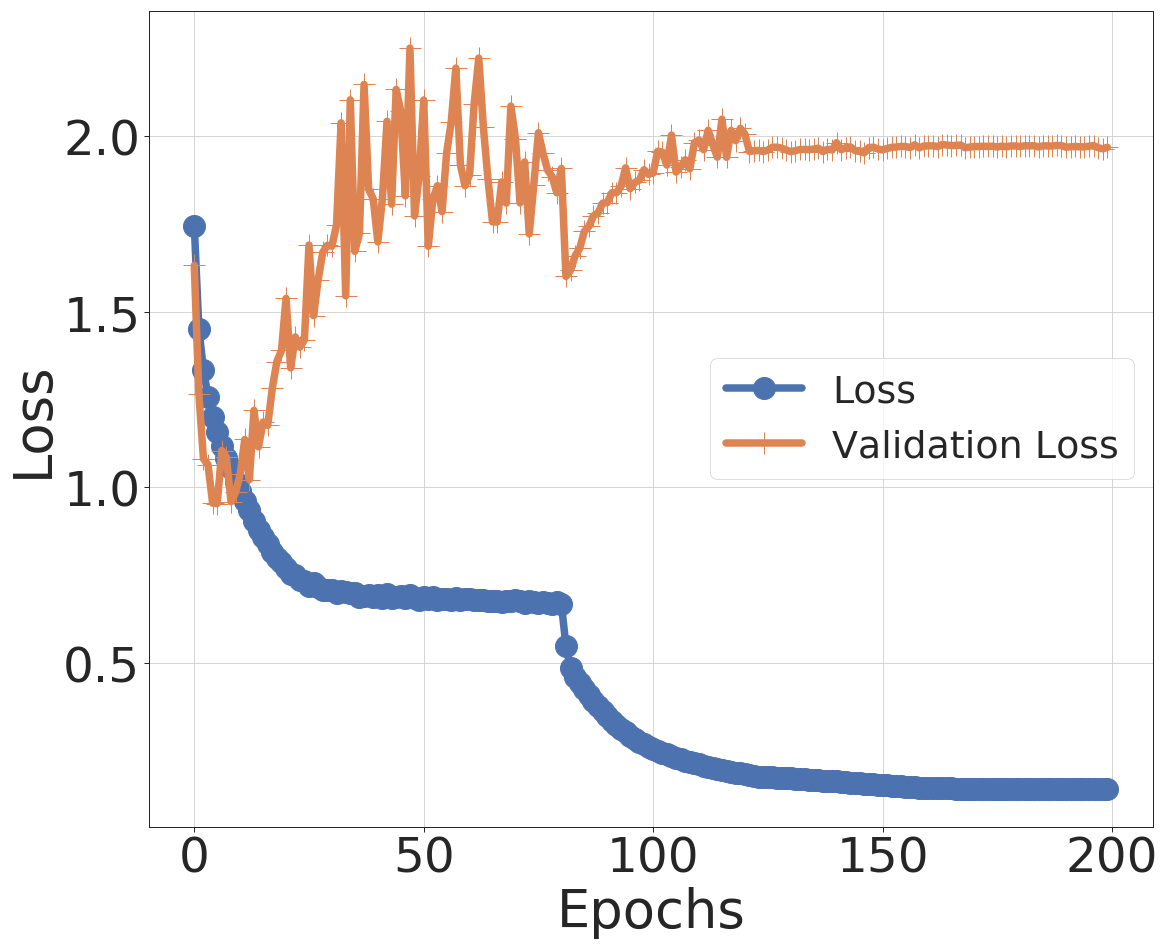}
	\caption{Loss over the epochs of a ResNet32, trained on noisy CIFAR10 data, with a fraction of $0.1$ erroneous labels.}\label{fig:resnet_loss}
	\vspace{-2.5mm}
\end{figure}

\begin{figure}[tbp]
	\centering
	\centering
	\includegraphics[width=0.99\linewidth]{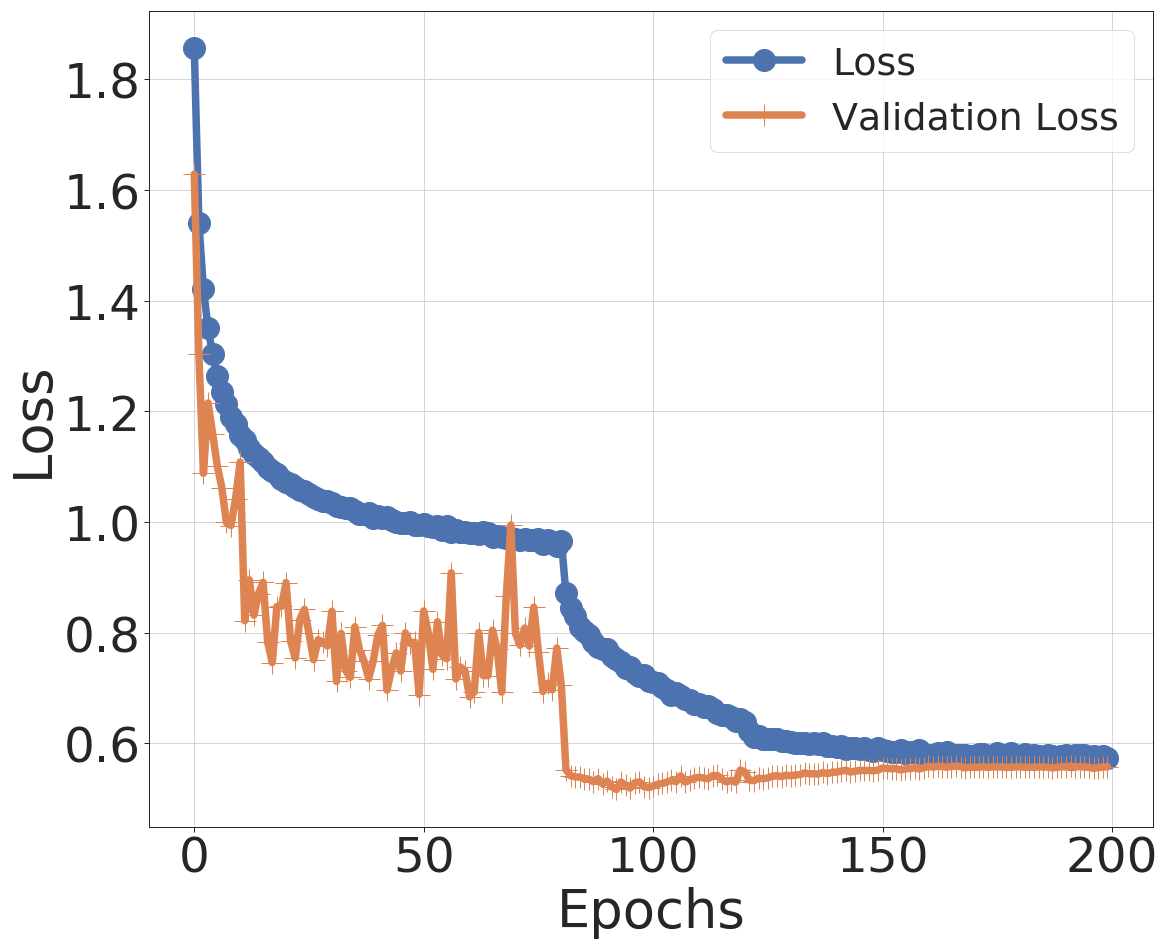}
	\caption{Loss over the epochs of a ResNet32, trained on noisy CIFAR10 data, with a fraction of $0.1$ erroneous labels, with data augmentation, which prevents overfitting the noisy labels.}\label{fig:resnet_loss_augmentation}
	\vspace{-2.5mm}
\end{figure}

As introduced in section V-A2 
 in the main document, high capacity networks such as ResNet32, can easily overfit the noisy training data. This critically hinders the success of methods dealing with erroneous labels. Which can not only be seen in the resulting performance of the optimisation, but also by monitoring the validation loss of a single training. Figure \ref{fig:resnet_loss} shows, that even for a noise fraction as small as $0.1$ the validation loss increases after approximately ten epochs, indicating an overfitting on the (in our case noisy) training data. We observe this even more prominently for higher noise fractions. Hence, regularisation is crucial for approaches as ours to work. As shown in Figure \ref{fig:resnet_loss_augmentation} the data augmentation is able to prevent overfitting the noisy data with the ResNet32 and the same setting as in Figure \ref{fig:resnet_loss}. This holds true for all other noise fractions investigated.

\section{Training details}
For all networks we stick as close as possible to the original setup (references given below). We neglect data augmentation, if not explicitly stated differently, to study the effect of changes to the labels alone (without any effect due to changes to the samples).\\

\textit{MNIST-CNN}: We use the architecture as defined in \cite{chollet2015keras}, using two convolutional layers and one dense layer, trained over 20 epochs with Adadelta. \\

\textit{CIFAR-CNN}: We use the four convolutional, one dense layer network as provided by \cite{chollet2015keras}, trained over 100 epochs, with RMSprop optimiser, a learning rate of $10^{-4}$ and a decay of $10^{-6}$.\\

\textit{ResNet}: We use a ResNet \cite{he2016deep} of depth 32, trained with the learning rate schedule as defined in \cite{chollet2015keras}, starting at $0.001$, changed to $0.0001$ at epoch 80, $10^{-5}$ at epoch 120, $10^{-6}$ at epoch 160, and $5\cdot 10^{-7}$ at epoch 180. The network is trained for 200 epochs. \\

\textit{Learning to reweight}: We use the code accompanying the publication \cite{ren2018learning}, uniform flip noise, and a ResNet of depth 32 \cite{he2016deep}. To compare the results directly to our approach we experiment with activating and deactivating data augmentation in the code, which was not done in the original paper by Ren et al.\\

\textit{Data augmentation}: We find data augmentation to play a key role in the performance of our approach and ``learning to reweight", if applied to a high capacity network, such as ResNet. To avoid the extreme overfitting to the noisy data we repeat experiments with ResNet including data augmentation, where we use horizontal flipping and padding the image with 4 additional pixels. This is the setup we found being used in the code accompanying \cite{ren2018learning}\footnote{\url{https://github.com/uber-research/learning-to-reweight-examples}} and reproduced for our experiments.

\end{document}